\documentclass{article}
\usepackage[T1]{fontenc}
\usepackage{arxiv}
\usepackage[authoryear,round]{natbib}
\setcitestyle{citesep={;},aysep={,},yysep={;}}
\renewcommand{\shorttitle}{From Search to Research}
\renewcommand{\headeright}{Preprint}
\renewcommand{\undertitle}{Preprint}
\date{}
\usepackage{graphicx,booktabs,longtable,array,amsmath,amssymb,float,pdflscape}
\usepackage{xcolor,fvextra}
\usepackage{hyperref}
\usepackage{url}
\hypersetup{hidelinks,
  pdftitle={From Search to Research: Exploring Search Scaling in Autonomous Quantitative Factor Mining},
  pdfauthor={Kangcheng Deng, Hui Cai, Jiacheng Lu, Chester Zhongshu Qian, Rui Sun, Beidi Luan, Jing Li, Daxin Jiang, Zuo Bai}}
\providecommand{\tightlist}{}
\definecolor{promptframe}{gray}{0.73}
\DefineVerbatimEnvironment{PromptBlock}{Verbatim}{fontsize=\footnotesize,breaklines=true,breakanywhere=true,breaksymbolleft={},frame=single,framesep=2mm,rulecolor=\color{promptframe}}

\title{From Search to Research:\\Exploring Search Scaling in Autonomous Quantitative Factor Mining}
\author{%
\parbox[t]{\dimexpr\textwidth-2\tabcolsep\relax}{\centering\normalfont
\textbf{Kangcheng Deng\textsuperscript{1}, Hui Cai\textsuperscript{1},
Jiacheng Lu\textsuperscript{1,2},
Chester Zhongshu Qian\textsuperscript{1,3}\thanks{Work done during internship at StepFun.},}\\
\textbf{Rui Sun\textsuperscript{1}, Beidi Luan\textsuperscript{1},
Jing Li\textsuperscript{1}, Daxin Jiang\textsuperscript{1},
Zuo Bai\textsuperscript{1,4,\ensuremath{\ddagger}}}\\[1.5ex]
\textsuperscript{1}StepFun, \textsuperscript{2}Shanghai Jiao Tong University\\
\textsuperscript{3}University of California, Los Angeles, \textsuperscript{4}FinStep\\[1.5ex]
\textsuperscript{\ensuremath{\ddagger}}Corresponding author.\\
Email: \href{mailto:baizuo@stepfun.com}{\texttt{baizuo@stepfun.com}}
}
}
\begin{document}
\raggedbottom
\maketitle

\begin{abstract}
Inference scaling has been shown to improve large language model (LLM) performance, and this principle naturally extends to autonomous LLM agents through increased search budgets, which we refer to as \emph{search scaling}. Although prior work has characterized the mechanisms, scaling behavior, and performance limits of LLM inference scaling, much less is known about these questions in autonomous research. Therefore, we investigate how search scaling affects research performance and what mechanisms drive these gains using 50 quantitative factor-mining tasks grounded in financial research reports. Each task requires an agent to carry out an end-to-end research loop, from interpreting a hypothesis and implementing it in code to evaluating and iteratively refining the resulting factor. Across nine models, we examine how model capability, search depth, and search organization shape factor quality by tracing performance across varying budgets, transferring intermediate research states between models, and comparing different search strategies. We find that (1) initial performance is more strongly associated with model capability, while deeper search can narrow cross-model gaps; (2) model grafting shows that the early research state materially shapes final performance; and (3) parallel search outperforms sequential search under the same iteration budget, consistent with benefits from broader coverage of the search space. Further trajectory analysis shows that higher-performing models more effectively diagnose failures, revise search directions, and preserve the intended economic hypothesis when selecting candidates. These findings suggest that future progress in autonomous research will require stronger models together with adaptive policies for deploying test-time computation throughout the research process.
\end{abstract}

\section{Introduction}\label{introduction}

Scaling parameters, data, and training compute has improved language-model capabilities \citep{kaplan2020scaling,hoffmann2022empirical}. Deployment-time scaling asks what additional computation can achieve once the model is fixed \citep{snell2025scaling}. When that computation is used to run dependent experiments, update an artifact, and carry empirical evidence into subsequent decisions, we call it \emph{search scaling}. This setting extends scaling beyond generating more tokens or independent candidate answers: an agent must observe results, diagnose failures, and choose the next experiment \citep{yao2023react,rank2026posttrainbench}. In this regime, added computation pays off only when the agent converts feedback into useful experiments, so the return to search may depend on the base model.

Existing test-time scaling studies examine independent sampling, verification, answer aggregation, and sequential answer refinement, including when smaller models benefit from larger inference budgets \citep{snell2025scaling,wu2025inference}. Research-engineering benchmarks also compare agents under different time budgets \citep{wijk2025rebench}. These results leave open how capability and budget interact in dependent empirical search, where agents choose experiments, interpret their results, and use that evidence to decide the next action. More experiments provide opportunities for improvement but also for mistaken diagnoses, unproductive local search, and adaptive selection that favors noise. Greater search depth therefore need not improve search quality. Initial quality and the ability to improve through feedback may also differ: a strong initial proposal need not imply sustained gains, while a weaker starting point may leave room for effective revision. The key question is not simply whether more research helps, but how base-model capability shapes the marginal value of additional experimentation and the performance level that iterative search can reach.

Quantitative factor construction provides a controlled yet non-trivial testbed for this question. Grounded in the search for signals that rank assets \citep{zhang2020autoalpha,yu2023generating}, our tasks require agents to construct factors without training a predictive model. Additional iterations therefore change research behavior without also increasing model-training compute, optimizer steps, or training stochasticity. Economic hypotheses constrain the search space, and LLM-based idea-alignment checks (\hyperref[validity-and-alignment-audit]{Appendix C.3}) discourage revisions that abandon the intended hypothesis to pursue backtest scores. These constraints limit unrestricted fitting, while repeated candidate selection on development backtests still creates multiple-testing risk. Noisy financial feedback, local optima, and adaptive selection preserve a substantive generalization problem, making held-out evaluation essential. With a fixed environment and shared harness, this setting lets us examine how agents use additional experimental opportunities to improve out-of-sample factor quality while preserving the intended economic hypothesis.

To answer these questions, we build a benchmark of 50 tasks grounded in quantitative research reports from the Chinese A-share market. Each task supplies an economic hypothesis and a factor-construction idea, and an agent working through a shared general-purpose harness implements the factor, runs development-period backtests, and revises its candidate; with the execution architecture fixed, the base model and the search budget are the principal experimental variables. We evaluate nine models along continuous ten-iteration trajectories: the harness freezes a checkpoint after every iteration, an independent evaluator scores it on a held-out period, and we report results at iterations 3, 6, and 10 with realized resource consumption recorded separately from nominal depth. Treating base-model capability, nominal search depth, and realized resource consumption as distinct quantities (\hyperref[problem-formulation]{Section 3.1}), we further conduct two controlled interventions, model grafting and a sequential--parallel comparison, to probe what makes iterative search effective. Our contributions are:

\begin{itemize}
\tightlist
\item
  We establish a training-free, hypothesis-constrained protocol for measuring search scaling through continuous research trajectories, fixed checkpoints, and held-out factor evaluation.
\item
  We characterize cross-model budget-performance curves and fit an empirical search scaling law, finding that base capability tracks initial quality more closely than marginal search gains and that deeper search can narrow some cross-model gaps.
\item
  We analyze cost and compute allocation through model grafting and sequential--parallel interventions, showing that early research-state quality and search breadth can matter as much as additional depth.
\item
  We connect trajectory behavior to outcome differences, finding that higher-performing models redirect search at the representation level, attribute feedback through controlled comparisons, and retain candidates that remain consistent with the assigned economic hypothesis.
\end{itemize}

\section{Related Work}\label{related-work}

\textbf{Scaling Laws and Test-Time Compute.} Training scaling laws relate language-model loss to parameter count, data volume, and training compute, guiding the allocation of pretraining resources \citep{kaplan2020scaling,hoffmann2022empirical}. Incorporating inference demand extends this allocation problem to deployment and can favor smaller models trained on more data \citep{sardana2024beyond}. Test-time scaling then treats computation per problem as a decision variable. Verifier-guided search, sequential revision, and comparisons of sampling, voting, tree search, and diversified agent rollouts already establish that smaller models can outperform larger ones with sufficient inference computation under suitable conditions \citep{snell2025scaling,setlur2025scaling,wu2025inference,zhu2025scalingagents}. General AgentBench extends sequential and parallel scaling to general agents and finds that added computation can be constrained by a context ceiling and a verification gap \citep{li2026generalagentscaling}. Model--compute substitution is therefore a starting point for studying research agents.

Connecting these results to autonomous research requires distinguishing compute allocation from feedback generation. Independent sampling explores alternative candidates, whereas sequential revision conditions later proposals on previous attempts. In the reasoning settings above, answer aggregation and learned outcome or process verification guide selection. Tool-using research agents additionally generate evidence by implementing hypotheses and executing experiments \citep{yao2023react,rank2026posttrainbench}. Although a research evaluator also serves a verification role, agents must decide which experiment to run and how its results should change the artifact. Research-time scaling thus depends on converting experimental feedback into useful revisions; in factor research, their value must also be assessed beyond the historical data that guided development.

\textbf{Autonomous Research and Experimentation.} Autonomous research spans full-cycle systems and evaluation environments. The AI Scientist links idea generation, implementation, experiments, and manuscript production, while MLE-bench evaluates competition-style machine-learning engineering and PaperBench tests research replication \citep{lu2024aiscientist,chan2025mlebench,starace2025paperbench}. Performance-driven benchmarks leave the solution open: RE-Bench provides research-engineering objectives with task-specific evaluation, ResearchGym supplies datasets, baselines, and evaluation code for discovering improved methods, and PostTrainBench optimizes a post-training pipeline evaluated on held-out benchmarks \citep{wijk2025rebench,garikaparthi2026researchgym,rank2026posttrainbench}. Across these settings, evaluation measures what an agent system accomplishes through sustained experimental work.

Budget allocation determines how that work can accumulate. RE-Bench varies both attempt duration and the number of independent attempts, combining longer trajectories with best-of-\(k\) selection. ResearchGym examines continuation by allocating additional resources to selected runs, and PostTrainBench reports time-budget curves, resource use, and scaffold comparisons \citep{wijk2025rebench,garikaparthi2026researchgym,rank2026posttrainbench}. Independent restarts explore alternative initial plans; continuation lets decisions build on accumulated code and empirical evidence. We study the latter mechanism by varying research depth within a single trajectory and examining how its returns depend on the base model.

\textbf{Trading Agents and Quantitative Factor Research.} Financial-agent evaluation connects decisions to their consequences in an evolving environment. Sequential investment evaluation in InvestorBench and StockBench has expanded toward live-market evaluation in DeepFund, LiveTradeBench, and AI-Trader \citep{li2025investorbench,chen2025stockbench,li2025time,yu2025livetradebench,fan2025aitrader}. Capability attribution and process diagnostics further ask what observed returns reveal about agent behavior \citep{zhu2026knowing,qu2026clqt}. OpenFinGym broadens this setting to executable tasks derived from financial publications, with verifiers and controls separating agent-accessible information from evaluation outcomes \citep{zhang2026openfingym}.

Factor research gives agents control over the predictive artifact itself, including its hypothesis, implementation, and revision. Search has expanded from mathematical expressions to executable programs and iterative procedures informed by accumulated experience \citep{yu2023generating,li2025rdagent,liu2026cognitive,wang2026factorminer}. Evaluation consequently covers iterative search and its cost in AlphaBench, executable strategy construction in AlphaForgeBench, and financial, semantic, and structural quality in AlphaEval and AlphaQT-Bench \citep{luo2026alphabench,zhang2026alphaforgebench,ding2026alphaeval,luo2026alphaqt}. AlphaBench already preserves evaluation history and the best factor across rounds in its chain-of-experience search. Building on these iterative research settings, our focus is how base-model capability changes the marginal value of additional sequential experimentation. Our work crosses models with research budgets to examine whether deeper search improves held-out factor quality, closes capability gaps, and offers favorable trade-offs in realized cost.

\section{Search Scaling in Autonomous Quantitative Factor Mining}\label{search-scaling-in-autonomous-quantitative-factor-mining}

\subsection{Problem Formulation}\label{problem-formulation}

Let \(x_i\) denote an autonomous factor-research objective, \(M_m\) a fixed base model, and \(H\) a shared general-purpose agent harness. The system follows one continuous trajectory \(\tau_{m,i}\) on task \(i\). At search depth \(b\), it selects an incumbent factor \(f_{m,i}^{(b)}\) using the evidence available during development. An independent evaluator measures its held-out quality as \(Q_{m,i}(b)=q(f_{m,i}^{(b)},D_{\mathrm{test}})\). The checkpoint is an artifact produced by research, whereas \(Q_{m,i}(b)\) is observed only by the evaluator.

We distinguish base-model capability \(C_m\), nominal search budget \(b\), and realized resource consumption \(R_{m,i}(b)\). The budget counts externally controlled research iterations. Resource consumption records the tokens, tool calls, elapsed time, and API cost incurred by depth \(b\); one iteration can contain different amounts of work across systems. Holding \(M_m\), \(H\), and \(x_i\) fixed, we ask how continuing the trajectory changes held-out quality. Across models, we ask whether capability is associated with different marginal returns and performance levels over the observed budget range.

\subsection{Controlled Iterative Search}\label{controlled-iterative-search}

An external harness loop advances each model--task pair along a single research trajectory. Each loop invocation permits one research cycle in which the agent can act on its current state, run experiments, inspect development-period feedback, and update its candidate. The next cycle inherits the workspace artifacts, memory, and interaction history. The agent is not told the eventual trajectory length. We save the incumbent after every iteration, so later checkpoints extend the same research process rather than restart it with a different budget instruction.

Checkpoint evaluation is separate from search. The evaluator measures a frozen incumbent on held-out data, and these results never enter later agent observations. This makes the change between successive checkpoints attributable to continued development-period research under the shared protocol. \hyperref[experiments]{Section 4} specifies the checkpoint depths, tasks, financial environment, and models; the prompts and state-handling details appear in \hyperref[experimental-configurations]{Appendix D}.

\hyperref[fig:evaluation-pipeline]{Figure 1} summarizes the development, delivery, and independent-evaluation stages of one research trajectory.

\begin{figure}[H]
\centering
\includegraphics[width=\linewidth]{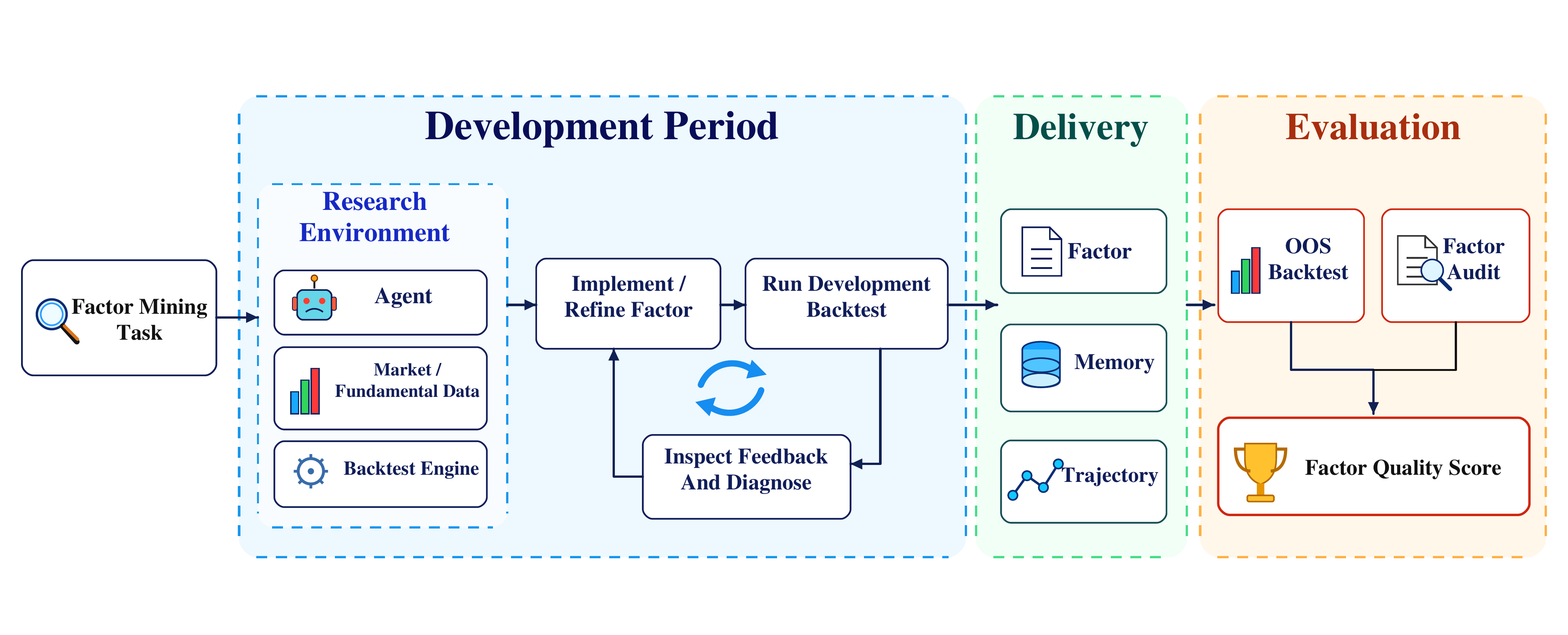}
\caption{Evaluation pipeline for autonomous factor research. During development, the agent implements factors, runs backtests, and revises its approach from empirical feedback. After every iteration, the harness freezes the current factor and retains its memory and trajectory. The evaluator independently combines an out-of-sample backtest with a factor audit to produce the quality score; evaluation feedback does not return to the agent.}
\label{fig:evaluation-pipeline}
\end{figure}

\subsection{Measuring Search Scaling}\label{measuring-search-scaling}

Search scaling is observed through the checkpoint outcomes produced as a trajectory grows deeper.

For a specified task-level quality measure, we aggregate held-out outcomes over the same \(N\) tasks at each depth: \[
\bar Q_m(b)=\frac{1}{N}\sum_{i=1}^{N}Q_{m,i}(b),
\qquad
\Delta_m(b_1,b_2)=\bar Q_m(b_2)-\bar Q_m(b_1),\quad b_2>b_1.
\] The paired increment \(\Delta_m\) measures the realized gain from continuing search over the observed interval. We record resource use at each checkpoint to compare quality with the actual cost of reaching it. \hyperref[experiments]{Section 4} specifies the task-level measure and the model-level score.

To describe the budget-performance curve, one candidate is a shifted, saturating power law: \[
\bar Q_m(b)=Q_{m,\infty}-A_m(b+b_0)^{-\alpha_m},\qquad A_m>0,\ \alpha_m>0.
\] In this family, \(Q_{m,\infty}\) is the plateau parameter, \(A_m\) sets the scale of remaining gains, \(\alpha_m\) controls their decay with depth, and \(b_0\) is a shared offset fixed before fitting. Power-law-type coverage under repeated sampling \citep{brown2024large} and saturation with larger inference budgets \citep{wu2025inference} motivate examining diminishing returns in autonomous research. We characterize search scaling through observed \(\bar Q_m(b)\), paired increments, and realized costs, then compare these quantities across levels of external model capability \(C_m\). The candidate curve provides an interpretable description of diminishing gains when the observed response follows that pattern; \hyperref[an-empirical-search-scaling-law]{Appendix D.4} fits this family jointly with the capability index.

\section{Experiments}\label{experiments}

\subsection{Experimental Setup}\label{experimental-setup}

We evaluate 50 report-grounded factor-research tasks spanning fundamental and technical ideas, including momentum, reversal, valuation, earnings, growth, volatility, and liquidity signals. All systems use the same Chinese A-share environment. Factor development uses data from January 1, 2013 through June 30, 2024; the evaluation period is July 1, 2024 through July 17, 2026. Source selection, data timing, and portfolio rules are detailed in \hyperref[benchmark-construction]{Appendix A} and \hyperref[research-environment-and-backtesting]{Appendix B}.

The main experiment comprises comparing model performance across nine models: Claude Opus 5 \citep{anthropic2026opus}, DeepSeek-V4.1-Flash \citep{deepseek2026v41flash}, DeepSeek-V4-Pro \citep{deepseekai2026v4}, Doubao Seed 2.1 Pro, GLM-5.2 and GLM-5.3 \citep{glm5team2026glm5}, MiniMax-M3 \citep{lai2026minimax}, Qwen3.7-Plus \citep{alibabacloud2026modelcatalog}, and DeepSeek-V3.1 \citep{deepseek2025v31}. We use Claude Code (v2.1.209) as the agent harness, and we launch it in bare mode to avoid the undesired impact of extensions, skills, or persistent memory on model behavior. For each model--task pair, the harness runs one continuous 10-iteration trajectory and saves a checkpoint after every iteration. The agent receives development feedback, while checkpoints are evaluated independently outside the agent workspace. Each model is evaluated five times on the 50 tasks, and the reported final score is the arithmetic mean of the five run-level scores, reducing sensitivity to a single stochastic trajectory. Runtime limits and state handling appear in \hyperref[experimental-configurations]{Appendix D}; the fixed mining and audit prompts are reported in \hyperref[factor-mining-prompt]{Appendix D.2} and \hyperref[validity-and-alignment-audit-prompt]{Appendix C.4}.

Each submitted factor is evaluated on three complementary aspects of quality. Net Sharpe ratio measures the risk-adjusted performance of the long--short portfolio constructed from the factor, monotonicity measures how consistently factor-sorted portfolios are ordered by the signal, and idea alignment measures the fidelity of the implementation to the assigned economic hypothesis. The task-level metrics are aggregated into the model-level score; GPT-5.6 Sol audits temporal validity and alignment independently of the tested models, and the scoring rubric is in \hyperref[metric-definitions-and-scoring]{Appendix C}. Actual API expenditure provides cost per task.

\subsection{Search Scaling across Models and Budgets}\label{search-scaling-across-models-and-budgets}

All models improve from three to ten iterations, but the returns to additional search differ substantially. \hyperref[tab:overall]{Table 1} reports the nine models at their recorded checkpoints. Claude Opus 5 rises from 75.45 to 92.50, DeepSeek-V4.1-Flash from 65.35 to 83.44, and MiniMax-M3 from 37.35 to 55.99. Their paths differ: Claude Opus 5 approaches its ten-iteration result by iteration 6, whereas DeepSeek-V4.1-Flash and MiniMax-M3 continue to gain later. DeepSeek-V3.1 improves sharply by iteration 6 and declines slightly by iteration 10.

\begin{table}[H]
\caption{Results across research depths.}
\label{tab:overall}
\centering
\scriptsize
\setlength{\tabcolsep}{1.2pt}
\renewcommand{\arraystretch}{1.08}
\resizebox{\linewidth}{!}{%
\begin{tabular}{@{}l*{15}{r}@{}}
\toprule
 & \multicolumn{3}{c}{Score} & \multicolumn{3}{c}{Sharpe} & \multicolumn{3}{c}{Monotonicity} & \multicolumn{3}{c}{Alignment} & \multicolumn{3}{c}{Cost per task (RMB)} \\
\cmidrule(lr){2-4} \cmidrule(lr){5-7} \cmidrule(lr){8-10} \cmidrule(lr){11-13} \cmidrule(lr){14-16}
Model & b=3 & b=6 & b=10 & b=3 & b=6 & b=10 & b=3 & b=6 & b=10 & b=3 & b=6 & b=10 & b=3 & b=6 & b=10 \\
\midrule
Claude Opus 5 & 75.45 & 87.69 & 92.50 & 2.008 & 2.334 & 2.661 & 0.952 & 0.964 & 0.989 & 4.04 & 4.04 & 3.70 & 53.03 & 68.71 & 116.94 \\
GLM-5.3 & 67.15 & 83.12 & 92.65 & 1.787 & 2.212 & 2.568 & 0.948 & 0.970 & 0.978 & 4.24 & 4.24 & 3.84 & 17.14 & 25.04 & 39.61 \\
DeepSeek-V4.1-Flash & 65.35 & 73.51 & 83.44 & 1.739 & 1.956 & 2.266 & 0.942 & 0.969 & 0.983 & 4.32 & 4.32 & 3.92 & 1.43 & 2.51 & 3.25 \\
DeepSeek-V4-Pro & 53.68 & 60.59 & 71.40 & 1.429 & 1.612 & 2.054 & 0.940 & 0.965 & 0.963 & 4.20 & 4.20 & 3.70 & 17.70 & 31.85 & 56.37 \\
Doubao Seed 2.1 Pro & 49.58 & 59.95 & 67.97 & 1.319 & 1.595 & 1.827 & 0.953 & 0.961 & 0.967 & 4.10 & 4.10 & 3.96 & 7.36 & 9.29 & 14.31 \\
GLM-5.2 & 46.84 & 58.73 & 60.29 & 1.246 & 1.563 & 1.707 & 0.945 & 0.960 & 0.959 & 4.32 & 4.32 & 3.76 & 10.39 & 14.94 & 26.89 \\
MiniMax-M3 & 37.35 & 45.27 & 55.99 & 0.994 & 1.309 & 1.611 & 0.902 & 0.953 & 0.960 & 4.20 & 3.68 & 3.70 & 25.22 & 41.89 & 55.86 \\
Qwen3.7-Plus & 34.30 & 46.65 & 51.06 & 0.976 & 1.328 & 1.709 & 0.904 & 0.932 & 0.946 & 3.74 & 3.74 & 3.18 & 7.44 & 8.85 & 13.39 \\
DeepSeek-V3.1 & 13.89 & 33.54 & 33.46 & 0.666 & 1.174 & 1.046 & 0.763 & 0.915 & 0.886 & 2.62 & 3.04 & 3.46 & 5.95 & 10.34 & 13.02 \\
\bottomrule
\end{tabular}
}
\end{table}

Because the model-level score uses the LLM auditor's alignment ratings, we validate this component against human judgments. Human quantitative experts independently produced a reference idea-alignment score for each submitted factor using the same 0--5 ordinal rubric. Agreement between these human ratings and the LLM auditor reached a linearly weighted Cohen's \(\kappa_w=0.71\), which gives progressively less credit as ratings move farther apart \citep{cohen1968weighted,cicchetti1971reliability}.

\begin{figure}[tbp]
\centering
\includegraphics[width=\linewidth]{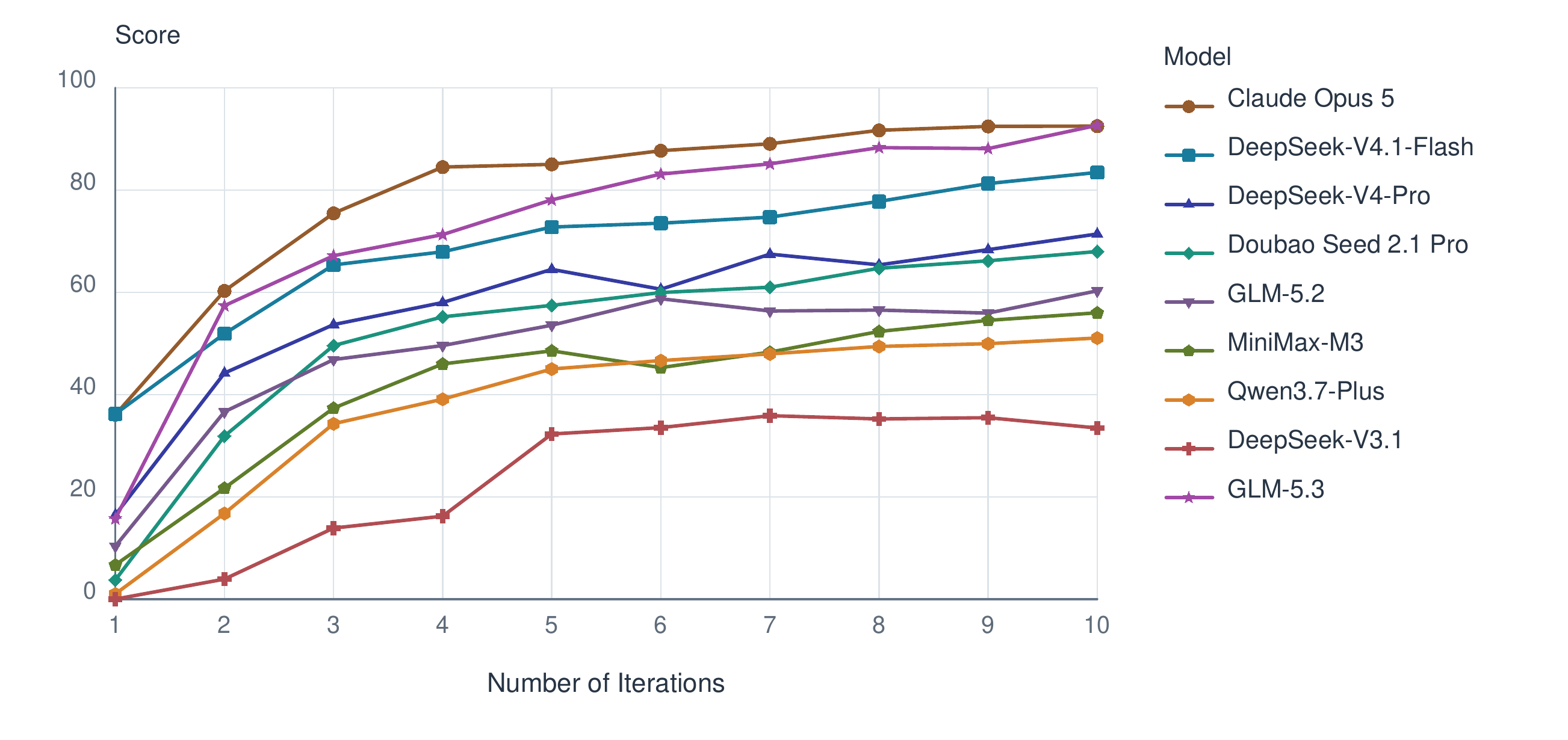}
\caption{Score across research iterations for nine models. Lines connect consecutive checkpoints; colors and marker shapes identify models.}
\label{fig:search-scaling}
\end{figure}

\hyperref[fig:search-scaling]{Figure 2} expands the budget view to iterations 1--10. The curves show that extra search can improve a weaker initial configuration substantially, while a strong initial configuration can approach a plateau earlier. The same nominal iteration increment does not produce the same score gain across models.

We use the September 25, 2026 snapshot of the Artificial Analysis Intelligence Index (AAII) v4.3.2 \citep{artificialanalysis2026intelligence} as the external capability proxy \(C_m\); \hyperref[external-capability-indicator]{Appendix D.3} records the exact model release and reasoning effort for each entry. Doubao Seed 2.1 Pro has no matching entry, so capability-based analyses use the other eight models. Their rank correlation between capability and three-iteration mean Sharpe is \(\rho_{\mathrm{Spearman}}=1.00\). The correlation with the three-to-ten-iteration Sharpe gain is \(0.48\). Model capability therefore closely tracks the starting level, while the gain from continued research follows a different ordering. Given additional budget, a relatively weaker model can reach or even exceed the level of a stronger one: DeepSeek-V4.1-Flash at ten iterations scores 83.44, exceeding Claude Opus 5 at three iterations (75.45).

\begin{figure}[H]
\centering
\includegraphics[width=\linewidth]{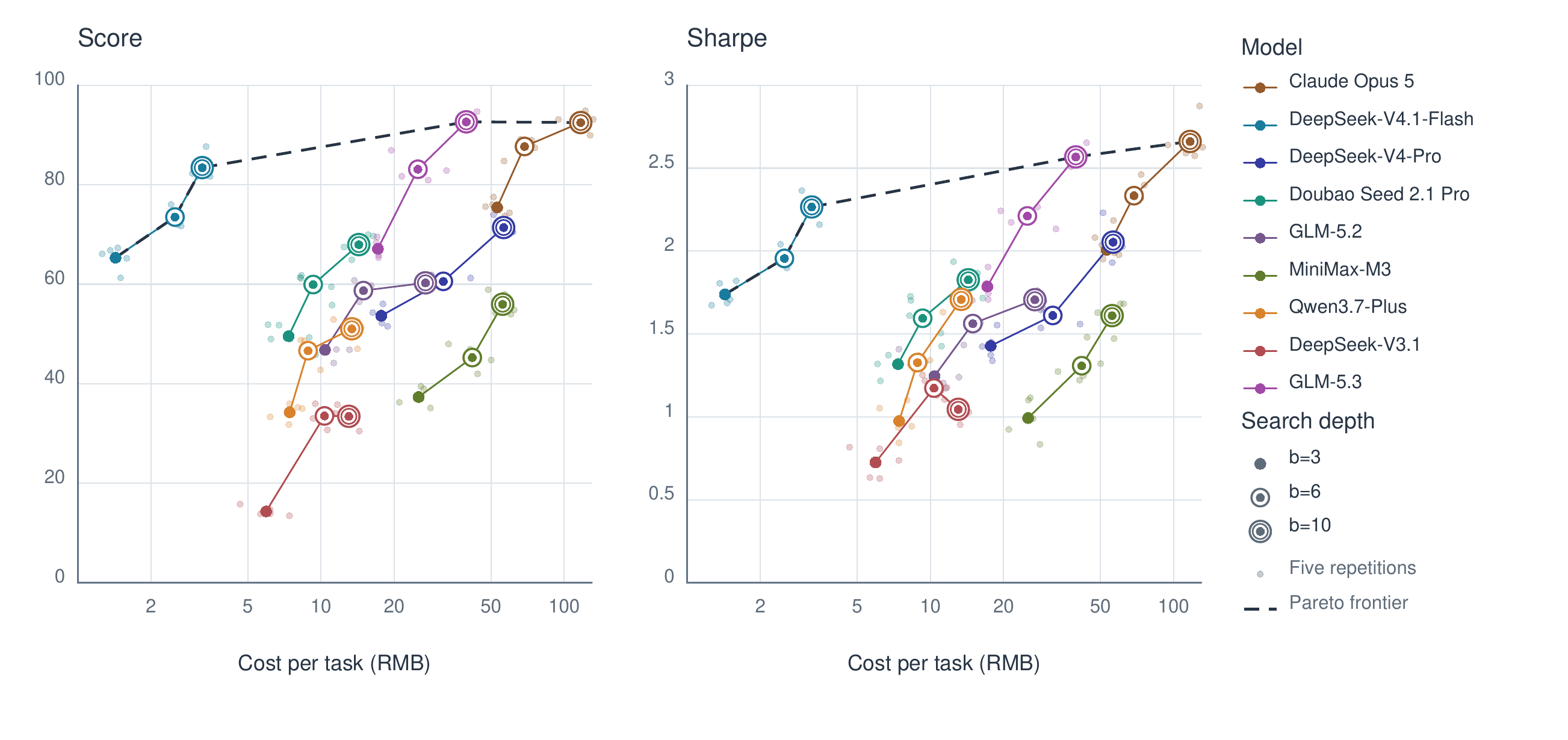}
\caption{Cost--performance relationships across search depths. The left panel plots Score and the right panel plots mean Sharpe against cost per task for $b=3,6,10$. Faint points show the five repetitions of each configuration, while the larger marker shows their mean. Colors identify models, marker rings encode search depth, solid paths connect checkpoints from the same model, and dark dashed curves trace the Pareto frontier. The horizontal axes use a log scale.}
\label{fig:cost-performance}
\end{figure}

Actual expenditure sharpens this comparison. \hyperref[fig:cost-performance]{Figure 3} places all 27 model--budget configurations at \(b\in\{3,6,10\}\) in cost--performance space; the left panel uses the model-level score, and the right panel isolates mean Sharpe. DeepSeek-V4.1-Flash dominates the low-cost end of the Pareto frontier in both panels: at ten iterations it reaches a score of 83.44 and a mean Sharpe of 2.266 for 3.25 RMB per task, roughly an order of magnitude cheaper than any other configuration of comparable quality. The high-quality end differs between the two panels: Claude Opus 5 at ten iterations attains the highest mean Sharpe of all configurations, but its weaker alignment ratings pull its score below GLM-5.3's ten-iteration configuration at three times the cost, leaving GLM-5.3 as the only high-quality frontier point in the score panel.

\subsection{What Makes Iterative Search Effective?}\label{what-makes-iterative-search-effective}

Section 4.2 shows that additional depth improves every model, but leaves open what makes iterative search effective. We therefore conduct two controlled experiments: model grafting separates the value of an early research state from the capability of the model that continues it, and a sequential--parallel comparison tests whether spending the same total budget on several independent searches outperforms deepening a single trajectory. In the grafting experiment, we transfer the accumulated factor, development evidence, and research memory between GLM-5.3 and Qwen3.7-Plus at iteration 3, then let the receiving model continue for seven further iterations under the ten-iteration total budget. \hyperref[tab:model-grafting]{Table 2} reports both transfer directions.

\begin{table}[htbp]
\caption{Model grafting under a ten-iteration total budget.}
\label{tab:model-grafting}
\centering
\small
\setlength{\tabcolsep}{3pt}
\begin{tabular}{lrrrrrr}
\toprule
Research schedule & Score & Sharpe & Monotonicity & Alignment & Turnover & IR \\
\midrule
Qwen3.7-Plus (3) $\rightarrow$ GLM-5.3 (7) & 72.28 & 2.161 & 0.976 & 3.56 & 8.05\% & 2.125 \\
GLM-5.3 (3) $\rightarrow$ Qwen3.7-Plus (7) & 76.41 & 2.210 & 0.970 & 3.68 & 7.26\% & 2.180 \\
\bottomrule
\end{tabular}
\end{table}

Starting from GLM-5.3 and continuing with Qwen3.7-Plus yields a 4.13-point higher score and a 0.049 higher average Sharpe than the reverse schedule; the two configurations use the same model pair and total search depth and differ only in which model constructs the first three iterations of research state. Their final gap shows that the quality of this early state considerably affects later performance: a stronger continuation model does not fully erase an initially weaker trajectory, while a weaker continuation model can still build on a stronger initial state.

The second experiment varies how the total budget is spent. For DeepSeek-V4.1-Flash, Doubao Seed 2.1 Pro, and MiniMax-M3, we compare one sequential ten-iteration trajectory with five independent two-iteration searches under the same iteration budget, selecting the parallel candidate using development feedback; \hyperref[tab:parallel-search]{Table 3} reports the comparison. Parallel search improves both score and Sharpe for all three models.

\begin{table}[htbp]
\caption{Sequential and parallel search under a ten-iteration total budget.}
\label{tab:parallel-search}
\centering
\small
\setlength{\tabcolsep}{8pt}
\begin{tabular}{@{}lrrrr@{}}
\toprule
 & \multicolumn{2}{c}{Score} & \multicolumn{2}{c}{Sharpe} \\
\cmidrule(lr){2-3} \cmidrule(lr){4-5}
Model & Parallel & Sequential & Parallel & Sequential \\
\midrule
DeepSeek-V4.1-Flash & 89.80 & 83.44 & 2.390 & 2.266 \\
Doubao Seed 2.1 Pro & 73.00 & 67.97 & 1.943 & 1.827 \\
MiniMax-M3 & 61.57 & 55.99 & 1.638 & 1.611 \\
\bottomrule
\end{tabular}
\end{table}

The consistent gain indicates that exploring several initial directions can be more valuable than repeatedly improving one locally chosen direction. Together, the two experiments show that the return to additional search depends on both the inherited research state and whether the budget broadens or deepens the search.

\subsection{Qualitative Analysis of Research Trajectories}\label{qualitative-analysis-of-research-trajectories}

Comparing Claude Opus 5 and DeepSeek-V4.1-Flash with the lower-scoring MiniMax-M3 and DeepSeek-V3.1 on matched fundamental and technical tasks reveals differences in how agents redirect search. The higher-scoring models move from local parameter tuning to changes in measurement, component construction, or aggregation when diagnostics undermine the current approach. In lower-scoring trajectories, successive window, smoothing, and threshold changes can continue around the same representation, with disappointing results attributed to an economic ceiling. The distinction is whether a plateau changes the agent's account of the problem and the scope of its next experiment.

Feedback attribution also differs. Claude Opus 5 and DeepSeek-V4.1-Flash inspect component contributions, gross versus net returns, coverage, and implementation consistency before deciding which change to retain. Their records include corrections to earlier explanations after controlled comparisons contradict them. MiniMax-M3 and DeepSeek-V3.1 also attempt structural changes, but some trajectories revise several assumptions together or interpret sign reversals as evidence against the economic hypothesis before resolving construction errors. Such misattribution spends additional search on the wrong cause and can make a viable direction appear exhausted.

Candidate selection determines how much of the search translates into the final score. The higher-scoring model records include retaining earlier candidates after failed revisions and rejecting higher-Sharpe variants that remove an economically required adjustment. Lower-scoring records include adding a separate reversal signal or dropping hypothesis-defining terms while pursuing improved backtests. The behavioral gap therefore concerns diagnosis, redirection, and objective-consistent selection together. It helps explain why additional rounds do not erase performance differences: more experimentation is useful when the agent identifies the relevant defect and preserves what has already worked. \hyperref[research-cases]{Appendix E.2} provides paired research cases for these three behavioral differences.

\section{Discussion}\label{discussion}

\subsection{Search Scaling Is the Accumulation of Research State, Not Merely More Inference}\label{search-scaling-is-the-accumulation-of-research-state-not-merely-more-inference}

Grafting shows that progress can survive a change of model through the inherited implementation, evidence, and research memory. With model weights fixed, this state changes what subsequent decisions condition on, so an early implementation has downstream value through both its current performance and the experiments it makes informative. The advantage of parallel search further indicates that a useful starting direction can matter more than extending the history of one trajectory. Search-state quality therefore concerns whether accumulated evidence supports better decisions, rather than how much history is retained.

\subsection{Model Capability Is Stage-Dependent Rather Than a Single Scalar}\label{model-capability-is-stage-dependent-rather-than-a-single-scalar}

Qwen3.7-Plus benefits from GLM-5.3's initial state more than GLM-5.3 recovers from Qwen3.7-Plus's, suggesting that capability differences are concentrated partly in initialization. Interpreting a hypothesis and choosing its representation impose different demands from refining an established implementation. This explains how a model can be a useful continuation agent despite weaker independent performance. A low standalone score can conceal competent refinement, while a strong standalone model can still struggle to recover from a poorly initialized trajectory. Effective capability thus depends on the stage and inherited state. API price provides another distinct ordering, as the strong cost--performance position of DeepSeek-V4.1-Flash illustrates.

\subsection{Search Scaling as a Compute-Allocation Problem}\label{search-scaling-as-a-compute-allocation-problem}

Depth extends a trajectory, breadth explores alternative initial states, and routing assigns models to different stages. Parallel search and grafting show why these choices matter: early compute can change the productivity of later compute. A broader allocation objective is \[
\pi^*: (\text{research state},\text{remaining budget})\longrightarrow(\text{model},\text{search action}),
\] with actions including refinement, restart, and final selection. The aim is to maximize independently evaluated quality within the resource budget. Our interventions examine controlled choices in this space and indicate that a configuration's cost--performance position depends on its allocation rule as well as its model and total budget.

\subsection{When Should Research-Time Scaling Generalize?}\label{when-should-research-time-scaling-generalize}

Search scaling should transfer when tasks provide informative external feedback, persistent artifacts, incremental revision, and independent final evaluation. Factor research and code optimization meet these conditions; repeatable data analysis and scientific modeling may also do so. Domains with ambiguous evaluators, long feedback delays, or expensive physical experiments can exhibit different compute-allocation patterns.

\section{Conclusion}\label{conclusion}

We studied how search depth changes out-of-sample quality in autonomous quantitative factor mining and how its returns interact with model capability and search organization. Measurements along continuous trajectories separate nominal iteration budgets from realized resources, and cross-model curves with the empirical search scaling law in \hyperref[an-empirical-search-scaling-law]{Appendix D.4} show that starting quality and continued-search gains vary across models. Model grafting and sequential--parallel interventions reveal that early research states can retain value across model switches and that broader exploration can outperform extending one trajectory. Autonomous research scaling therefore depends on the amount of computation, the research state it creates, and how later computation is allocated across directions and stages.

\section{Limitations and Future Work}\label{limitations-and-future-work}

The capability--budget relationship and empirical search scaling law describe a finite set of models and search depths in a structured Chinese A-share factor-research environment. Longer horizons and later model generations require testing functional stability, parameter transfer, and prediction for unseen models and budgets. Broader model coverage and denser checkpoints can support this longitudinal analysis.

Our depth, breadth, and fixed model-grafting comparisons are controlled slices of a broader research-policy space. Future work can combine routing, continuation, restart, parallel exploration, and stopping in a state-dependent allocation policy, then test it in code optimization, data analysis, and scientific modeling tasks with executable feedback, persistent artifacts, and iterative revision.

\subsection*{AI use statement}
Generative AI tools assisted with research ideation, experimental design and implementation, formulating mathematical claims and developing supporting arguments, and analyzing research trajectories and experimental results. They also supported literature retrieval and citation checking, manuscript drafting, translation and language editing, and figure and table preparation. Claude Opus 5 assisted with report screening and generated the task descriptions used to construct the research-task dataset (\hyperref[benchmark-construction]{Appendix A}); human quantitative experts selected the reports and reviewed the generated objectives. A separate LLM auditor evaluated factor implementations as described in \hyperref[experimental-setup]{Section 4.1}. The authors reviewed the AI-assisted text, mathematical arguments, code, and analyses and take responsibility for the final content, claims, and artifacts.

\bibliography{references}
\bibliographystyle{plainnat}

\appendix
\numberwithin{table}{section}
\numberwithin{figure}{section}
\section{Benchmark Construction}\label{benchmark-construction}

\subsection{Report Selection and Task Composition}\label{report-selection-and-task-composition}

The source corpus consists of quantitative research reports issued by multiple sell-side brokerages between February 2017 and June 2024. We retained reports about a single factor whose main contribution was factor construction followed by a single-factor backtest, with an explicit construction formula and Chinese A-share stocks as the investment universe. Claude Opus 5 \citep{anthropic2026opus} performed an initial screening; human quantitative experts then reviewed the candidates and selected 50 reports. The final set contains 30 technical and 20 fundamental tasks, spanning daily and intraday inputs and multiple proposed sources of alpha. \hyperref[tab:technical-composition]{Table A.1} and \hyperref[tab:fundamental-composition]{Table A.2} summarize the two task groups by signal family and input frequency.

\begin{table}[htbp]
\caption{Technical tasks by signal family and input frequency.}
\label{tab:technical-composition}
\centering
\small
\setlength{\tabcolsep}{3pt}
\setlength{\tabcolsep}{6pt}
\renewcommand{\arraystretch}{1.18}
\begin{tabular}{@{}l*{5}{>{\centering\arraybackslash}p{0.48in}}>{\centering\arraybackslash}p{0.48in}>{\centering\arraybackslash}p{0.48in}>{\centering\arraybackslash}p{0.48in}>{\centering\arraybackslash}p{0.48in}>{\centering\arraybackslash}p{0.48in}>{\centering\arraybackslash}p{0.48in}>{\centering\arraybackslash}p{0.48in}>{\centering\arraybackslash}p{0.48in}>{\centering\arraybackslash}p{0.48in}>{\centering\arraybackslash}p{0.48in}@{}}
\toprule
Signal family & 1 min & 5 min & 30 min & Daily & Total \\
\midrule
Volatility & 2 & - & - & 3 & 5 \\
Momentum & 3 & - & 1 & 2 & 6 \\
Liquidity & 3 & - & - & 3 & 6 \\
Microstructure & 2 & 1 & - & 1 & 4 \\
Prospect theory & 1 & 3 & - & 2 & 6 \\
Beta & 2 & - & - & 1 & 3 \\
\midrule
\textbf{Total} & \textbf{13} & \textbf{4} & \textbf{1} & \textbf{12} & \textbf{30} \\
\bottomrule
\end{tabular}
\par\smallskip{\footnotesize Counts are tasks; short dashes denote zero.}
\end{table}

\begin{table}[htbp]
\caption{Fundamental tasks by signal family and input frequency.}
\label{tab:fundamental-composition}
\centering
\small
\setlength{\tabcolsep}{3pt}
\setlength{\tabcolsep}{6pt}
\renewcommand{\arraystretch}{1.18}
\begin{tabular}{@{}l*{4}{>{\centering\arraybackslash}p{0.70in}}>{\centering\arraybackslash}p{0.70in}>{\centering\arraybackslash}p{0.70in}>{\centering\arraybackslash}p{0.70in}>{\centering\arraybackslash}p{0.70in}>{\centering\arraybackslash}p{0.70in}>{\centering\arraybackslash}p{0.70in}>{\centering\arraybackslash}p{0.70in}>{\centering\arraybackslash}p{0.70in}@{}}
\toprule
Signal family & 5 min & Daily & \shortstack{Fundamentals\\only} & Total \\
\midrule
Valuation & - & 8 & - & 8 \\
Profitability & - & 4 & - & 4 \\
Earnings surprise & 1 & 1 & - & 2 \\
Growth & - & 1 & 1 & 2 \\
Dividends & - & 2 & - & 2 \\
Quality & - & 1 & - & 1 \\
Leverage & - & - & 1 & 1 \\
\midrule
\textbf{Total} & \textbf{1} & \textbf{17} & \textbf{2} & \textbf{20} \\
\bottomrule
\end{tabular}
\par\smallskip{\footnotesize ``Fundamentals only'' denotes tasks that do not require quote data for the core construction.}
\end{table}

The frequency label records the principal quote-data resolution associated with the task, rather than the eventual frequency of portfolio rebalancing. Signal families are descriptive labels for the originating research idea.

\subsection{Task-Generation Prompt and Human Review}\label{task-generation-prompt-and-human-review}

The task-generation instructions are given below. The placeholders \texttt{\{source\_block\}} and \texttt{\{data\_block\}} denote the source material and the available-data catalog supplied at execution time.

\begin{PromptBlock}
You are a quantitative factor researcher generating a factor-mining task for the benchmark. Do not write code. Produce a clear factor-research direction for a subsequent mining loop to implement and evaluate.

# Source for this idea
{source_block}

# Available data

The following data block is generated from `catalog.json` and is always current:

{data_block}

See `data_catalog.md` for field units, caveats, and point-in-time (PIT) rules.

The available-data information is for reference. You are not required to use only the listed data when generating an idea. The downstream mining loop will handle data availability; do not let it determine the idea.

# Requirements

1. First use Bash/Read to inspect the provided `factor.py`, `card.md`, and `metrics.json` examples and the candidate records to avoid duplicates.
2. If useful, search the web to understand the current research on related factors, such as how a factor family has performed in the A-share market.
3. Produce a **concrete, implementable** factor idea. Its core should be the **calculation logic or formula**: which fields to use, how to combine them into a signal, and what the signal is expected to mean for stock selection. Provide enough information to implement the logic; do not be vague.

   At the same time, avoid unnecessary implementation detail. A broad formula and its economic mechanism are generally enough; mention important pitfalls only if helpful. Do not specify details as narrowly as "revenue growth g_0 = S_0/S_{{-1}} - 1, clipped to [-0.5, 1.0]; clipping is mandatory because an extreme AR(1) starting value would destabilize the accounting path." Excessive detail restricts the later mining agent's choices.

   A good idea has the following style:
   ```
   Construct a turnover-weighted short-term reversal factor from daily close and turnover_ratio. Use approximately 20 trading days in the window t-21 to t-1, skipping the most recent day to reduce bid-ask-bounce noise. Compute each day's return r_d = pct_change and turnover to_d; set w_d = to_d / sum(to_d) and wret = sum(r_d * w_d), so days with high turnover receive more weight. Define the factor as -wret / (standard deviation of daily returns in the window + 1e-6). A high value indicates a stock sold sharply on high-turnover, high-attention days and therefore a potential rebound; a low value indicates a stock bid up on high-turnover days and therefore a potential reversal. Require at least ten valid days, remove inf/NaN values, and use the cross-sectional factor directly for sorting; direction=1. Turnover weighting distinguishes an overreaction-driven selloff, which may reverse, from a low-volume decline that may continue. It may also help avoid a weak long leg or poor monotonicity in the middle groups.
   ```
   This idea describes how to calculate the factor and briefly explains the economic mechanism without prescribing excessive implementation detail.

   An overly detailed idea has the following style:
   ```
   Define trailing-twelve-month quantities with quarter-specific branches and normalize them by average assets. Construct total accruals from net profit and operating cash flow, while prescribing statement-scope reconciliation, several fallback fields, missing-data rules, clipping thresholds, and separate adjustments for non-cash items. Continue with additional field-by-field exceptions and parameter choices: ...
   ```
   This example specifies too many incidental accounting and fallback choices.

4. **Describe only the factor itself.** Do not include the rebalancing schedule, trading costs, backtest interval, stock universe, or exclusion rules. The framework fixes these settings (including ST stocks, recently listed stocks, suspensions, and price-limit cases); including them in the idea adds noise and may conflict with the framework. The idea must be self-contained: do not mention any research report, paper, source factor, historical conclusion, comparison result, or external claim such as "the report shows" or "research finds."

# Output format

Finally, output one JSON object inside a ```json code block:

```json
{{
  "name": "A short English factor name used as `<name>.py`, such as earnings_surprise_mom",
  "idea": "Factor calculation logic: which data fields to use, how to turn them into a cross-sectional signal (a formula or clear steps), and what high and low values imply for stock selection. Use 3-6 sentences focused on the factor and its logic. Omit rebalancing, costs, universe rules, and other framework settings.",
  "rationale": "Why this direction might work (1-2 sentences about the market mechanism)",
  "data_needed": "Required datasets, such as daily, forecast, or income",
  "source": {{
    "type": "research_report | improve_factor | direction (matching the current source)",
    "ref": "Specific provenance: internal report identifier for a research report; factor name for an existing-factor improvement; or direction description for an open research direction",
    "excerpt": "Key support for the idea: relevant passages for a report; the weakness being addressed for an existing factor; or the central insight for an open direction (1-3 sentences)"
  }}
}}
```
**Fill the source field truthfully.** It records the factor's provenance in its `card.md` so the theoretical basis can be reviewed later.
\end{PromptBlock}

Human reviewers retained objectives that were self-contained and specific enough to identify the required financial quantities, construction logic, signal direction, and economic mechanism. They rejected descriptions that were too vague to implement or so prescriptive that they fixed incidental parameters and removed meaningful research choices. Reviewers also checked that each objective remained grounded in a selected single-factor report and concerned the factor itself rather than benchmark-level portfolio or backtest settings. The rationale and supporting excerpt supplied alongside each generated candidate supported this review; they are not part of the agent-facing research objective.

\section{Research Environment and Backtesting}\label{research-environment-and-backtesting}

\subsection{Universe, Data, and Information Timing}\label{universe-data-and-information-timing}

The evaluation universe comprises approximately 5,200 stocks listed on the Shanghai and Shenzhen exchanges. At each signal date, the engine excludes special-treatment (ST) stocks and stocks with fewer than 60 days of listing history. The training (factor-development) period is January 1, 2013 through June 30, 2024, and the held-out test period is July 1, 2024 through July 17, 2026. Because every source report was published by June 2024, the test period contains only market observations subsequent to the report corpus. This temporal separation prevents look-ahead from report-derived task information and makes the reported financial evaluation out of sample. Daily price and volume data, intraday bars, industry classifications, valuation data, forecasts, preliminary earnings disclosures, and the three principal financial statements are available through a shared catalog. These inputs allow agents to construct factors from different economic and market mechanisms within the same equity universe.

For a signal dated \(T\), the factor interface supplies daily observations only through \(T\). Fundamental records are selected according to their publication dates, so the agent receives records available by \(T\) rather than a subsequently revised full-history panel. The audit separately checks whether submitted code bypasses this interface or otherwise introduces information unavailable at the signal date. This separation is material for accounting tasks: a correctly written formula can still be invalid if it uses a later disclosure as though it were known earlier.

\subsection{Factor Artifact and Postprocessing}\label{factor-artifact-and-postprocessing}

An agent submits an executable cross-sectional factor and the signal matrix produced from it. It chooses historical windows, field combinations, treatment of missing and extreme values, and postprocessing appropriate to the hypothesis. The common default postprocessing winsorizes the raw cross-section, removes industry and size exposures, and centers ranks. Agents may replace the default specification; the submitted postprocessing is retained with the artifact so that the evaluated signal can be interpreted together with the raw construction.

The agent begins in a fresh workspace containing the data catalog, documentation, common tools, and fixed implementation examples. It does not receive the broader factor library or prior performance records. Thus a task describes a construction objective, while the agent must turn that description into an executable factor and diagnose the effects of its own implementation and processing choices.

\subsection{Portfolio and Re-evaluation Protocol}\label{portfolio-and-re-evaluation-protocol}

Signals formed at the close of trading day \(T\) determine positions entered at the open of \(T+1\) and closed at the open of \(T+2\). The main long--short portfolio assigns centered cross-sectional rank weights, normalized to zero net exposure and unit gross exposure. Daily turnover is the sum of absolute changes in those weights. Net returns deduct 5 basis points on purchases and 15 basis points on sales. The evaluator also sorts stocks into ten factor-ranked groups and records each group's annualized return for the monotonicity calculation.

The benchmark recomputes financial metrics from the submitted signal in the common evaluator rather than accepting a metric file produced inside the agent's workspace. Agents receive backtest feedback only from the training period. Once an artifact is selected, the evaluator applies it to the held-out test period without returning those results for further revision. This ensures that agent-side changes to scoring code cannot redefine the reported outcome and that test performance remains out of sample. The implementation audit uses the submitted code and objective to assess temporal validity and semantic alignment. Financial metrics and audit findings are retained separately so that a profitable but invalid or off-objective artifact remains inspectable.

\section{Metric Definitions and Scoring}\label{metric-definitions-and-scoring}

\subsection{Financial Metrics}\label{financial-metrics}

Let \(r_t^{\mathrm{net}}\) denote the daily net return of the rank-weighted long--short portfolio. With 252 trading days per year, its Sharpe ratio is \(S=\sqrt{252}\,\overline r^{\mathrm{net}}/\operatorname{sd}(r_t^{\mathrm{net}})\). Annualized return is the compounded return over the observed days, rescaled to 252 days; maximum drawdown is the largest peak-to-trough decline of the net-value series, including the initial value of one as a possible peak. Daily average turnover measures the mean absolute change in portfolio weights. The information ratio (IR) is the annualized mean of the long portfolio's eligible-universe-benchmark excess return divided by the annualized standard deviation of that excess return. The cross-sectional information coefficient (IC) uses Spearman correlation between signals and subsequent returns; the information coefficient information ratio (ICIR) summarizes the mean IC relative to its temporal variation.

For monotonicity, the evaluator computes the annualized returns \(R_1,\ldots,R_{10}\) of the ten factor-sorted groups and reports \(M=|\rho_{\mathrm{Spearman}}((1,\ldots,10),(R_1,\ldots,R_{10}))|\). Taking the absolute value accommodates either factor direction, after the submitted direction has been oriented so that higher evaluated values correspond to the long side. This statistic measures cross-sectional ordering; it does not account for the trading costs and signal persistence that affect net Sharpe ratio.

\subsection{Model-Level Score}\label{model-level-score}

For each model--budget configuration, let \(\overline{S}\) denote mean net Sharpe ratio, \(M\) mean monotonicity, and \(A\) mean alignment divided by five. The final score is \[
\mathrm{Score}=\max\!\left(0,\,100\,\frac{\overline{S}}{S_{\mathrm{ref}}}\,P_M(M)\,P_A(A)\right),
\] \[
P_M(M)=
\begin{cases}
1, & M\geq\tau_M,\\
M/\tau_M, & M<\tau_M,
\end{cases}
\qquad
P_A(A)=
\begin{cases}
1, & A\geq\tau_A,\\
A/\tau_A, & A<\tau_A.
\end{cases}
\] We set \(\tau_M=0.9\) and \(\tau_A=0.8\). The shared reference is \(S_{\mathrm{ref}}=2.6611\), the mean Sharpe across 50 tasks achieved by Claude Opus 5 after ten research iterations. It remains constant across models and iteration budgets.

Let \(a_i\) be the audit's 0--5 alignment rating for task \(i\). The adjusted inputs \(\widetilde{S}_i\), \(\widetilde{M}_i\), and \(\widetilde{a}_i\) equal the corresponding task metrics for valid submissions. All three are set to zero when the task fails the leakage audit, is assigned zero for a protocol violation, or fails to produce an evaluable factor. Thus, \[
\overline{S}=\frac{1}{50}\sum_{i=1}^{50}\widetilde{S}_i,\qquad
M=\frac{1}{50}\sum_{i=1}^{50}\widetilde{M}_i,\qquad
A=\frac{1}{50}\sum_{i=1}^{50}\frac{\widetilde{a}_i}{5}.
\] The denominators remain fixed even when some tasks receive zero inputs. We compute the resulting score separately for each of the five repetitions and report their arithmetic mean. We first aggregate the task metrics within a repetition and then apply the two quality factors. Each factor imposes a linear penalty below its threshold and equals one above it. The outer maximum sets the score floor at zero, while the score has no upper cap: with both factors equal to one, a mean Sharpe above \(S_{\mathrm{ref}}\) yields a score above 100. IR, annualized return, drawdown, turnover, IC, and ICIR remain diagnostic measures.

\subsection{Validity and Alignment Audit}\label{validity-and-alignment-audit}

A separate, read-only GPT-5.6 Sol session \citep{openai2026gpt56} reviews each final implementation. GPT-5.6 Sol is not included among the evaluated systems, preventing a tested model from rating its own output and reducing self-preference bias. The validity judgment asks whether the factor uses only data available through the signal date, including whether it accesses files or data-loading paths outside the time-bounded interface. A leakage failure or a protocol violation sets the task's Sharpe, monotonicity, and alignment inputs to zero while retaining the task in the aggregate denominator.

The alignment judgment considers both fidelity to the assigned research hypothesis and the economic interpretability of the resulting factor. It does not require a verbatim reproduction of the provided idea: changes that retain the hypothesis and have a coherent economic rationale are allowed. This criterion instead prevents historical backtest optimization from replacing the assigned hypothesis with unrelated predictors. A rating of 5 means the core single-factor idea is faithfully implemented without unrelated signals; 3--4 allows minor interpretable additions; 1--2 reflects substantial dilution by unrelated predictors; and 0 denotes little connection to the objective. The audit records a rationale and identifies any drift signals.

\subsection{Validity and Alignment Audit Prompt}\label{validity-and-alignment-audit-prompt}

The audit session instantiates \texttt{\{factor\_filename\}} and \texttt{\{idea\_block\}} for each submitted artifact.

\begin{PromptBlock}
You are a rigorous quantitative-factor code auditor. Review an A-share cross-sectional factor in read-only mode. Do not modify any file. Assess temporal validity and idea alignment, then return one structured JSON object.

# Factor under review
- Factor implementation: `{factor_filename}`
- Read this file first. If the data interface requires clarification, read `factor_base.py`, which defines the factor interface.

# Point-in-time guarantee provided by the framework

The point-in-time (PIT) boundary is enforced in `loader.py`. At each signal date T, the factor receives daily matrices truncated at T, fundamental and forecast records filtered by publication date, and the eligible universe at T. Historical calculations that remain inside this time-bounded interface, including rolling statistics and shifts to earlier observations, are temporally valid.

# Dimension A: future-information leakage

The central question is whether the implementation escapes the time-bounded interface. Review the code line by line and reason about the following indicators:
- File I/O such as `open`, `pd.read_parquet`, `pd.read_csv`, `glob`, or direct use of `os` or `pathlib` to load data.
- Imports outside `factor_base`, `numpy`, `pandas`, `scipy`, and `math`. Importing `loader`, `DataLoader`, or another data-access module escapes the interface.
- Direct references to `DataLoader`, loader objects, or full-history matrices.
- Absolute paths, parent-directory traversal, hard-coded dates, `datetime.now()`, or fixed timestamps that introduce information unavailable at T.
- Forward-looking operators such as negative shifts or indices aligned to future returns or prices. A negative shift within a truncated historical window may only create missing values, so determine whether the code actually accesses data after T.
- Any external state that would not be known at T.

Decision rules:
- If the factor uses only the time-bounded interface and the allowed libraries, with no effective access to future information, set `verdict` to `PASS`.
- If the code accesses information outside the time-bounded interface, set `verdict` to `FAIL`.
- Do not flag valid historical rolling calculations, `iloc[-1]`, or positive shifts.
- For a failure, explain the causal path to leakage and list each finding with its severity and line number.

# Dimension B: alignment with the assigned idea

{idea_block}

Assess whether the implementation preserves the assigned factor direction or combines it with unrelated sources of signal.

- Core criterion: a single-factor task should not mix the intended signal with independent predictors. For example, adding reversal, momentum, or turnover components to a profitability-quality idea without a hypothesis-consistent reason constitutes drift.
- Supporting evidence: economic interpretability and a coherent formula support alignment. Excessive complexity or loosely connected components may indicate drift, but complexity alone is insufficient for a low score.
- Score the implementation from 0 to 5:
  - 5: faithfully implements the single intended direction without unrelated signals;
  - 3-4: preserves the main hypothesis with minor, interpretable additions;
  - 1-2: unrelated signals substantially dilute the intended direction;
  - 0: has little connection to the idea or combines several unrelated signal sources.
- Explain the score. For scores of 3 or below, list each unrelated or drifting component in `drift_signals`.
- If no idea is supplied, set `score` to `null` and state that there is no reference idea.

# Output

Return only the following JSON object inside a `json` code block:

```json
{
  "leak": {
    "verdict": "PASS or FAIL",
    "findings": [
      {
        "severity": "high/medium/low",
        "line": "line number",
        "issue": "short description",
        "why": "why this does or does not create leakage"
      }
    ],
    "reasoning": "overall temporal-validity reasoning"
  },
  "idea_alignment": {
    "score": "integer from 0 to 5, or null",
    "reasoning": "reason for the score",
    "drift_signals": ["drifting component 1", "drifting component 2"]
  }
}
```
\end{PromptBlock}

\section{Experimental Configurations}\label{experimental-configurations}

\subsection{Systems and Common Settings}\label{systems-and-common-settings}

The experiments compare nine deployed models, all using Claude Code (v2.1.209) in bare mode. This prevents extensions such as skills and persistent context from global memory files from influencing model behavior. The models and their versions are listed in \hyperref[experimental-setup]{Section 4.1}, and the common settings are summarized in \hyperref[tab:settings]{Table D.1}. We report model-level outcomes because each model selects and interprets experiments through the shared execution harness.

\begin{table}[H]
\caption{Main experimental settings.}
\label{tab:settings}
\centering
\small
\setlength{\tabcolsep}{3pt}
\begin{tabular}{lp{0.61\linewidth}}
\toprule
Setting & Value \\
\midrule
Research tasks & 50 fixed objectives: 20 fundamental, 30 technical \\
Score aggregation & All 50 tasks; failed or invalid tasks contribute zero scoring inputs \\
Research iterations & One continuous 10-iteration trajectory per model--task pair; a checkpoint after every iteration \\
Independent repetitions & 5 complete runs per model; reported Scores average the 5 run-level Scores \\
Reasoning effort & High where supported \\
Session timeout & 8 hours \\
Individual backtest timeout & 45 minutes \\
Training period & January 1, 2013--June 30, 2024 \\
Held-out test period & July 1, 2024--July 17, 2026 \\
Portfolio formation & Signal at close on day $T$; entry at open on day $T+1$; exit at open on day $T+2$ \\
Trading cost & 5 bp on purchases; 15 bp on sales \\
Independent final audit & GPT-5.6 Sol \\
\bottomrule
\end{tabular}
\end{table}

The external harness loop invokes one research cycle at a time without revealing the trajectory endpoint. State persists across checkpoints, while held-out outcomes remain outside the agent's context. The clock limits guard against stalled execution.

\subsection{Factor-Mining Prompt}\label{factor-mining-prompt}

The common harness instantiates the following template with the task idea, runtime filenames, backtest limits, and data catalog. Each invocation advances one externally controlled research cycle while preserving the existing memory, artifacts, and transcript. The total trajectory length is not disclosed to the agent.

\begin{PromptBlock}
You are a quant researcher. Your objective is to do factor mining according to the idea provided. You should implement and test a candidate, interpret the evidence, update the research memory, and yield a delivery as your submission.

# Factor idea
{idea}

# Research cycle

1. Inspect the current state.
   - Read `{memory_file}` and the existing candidate files.
   - On the first invocation, read the provided `factor.py` examples to learn the required implementation interface.
2. Implement or revise the candidate in `factor.py`.
3. Run the common backtest:
   `python backtest.py --factor <name> --start {bt_start} --end {bt_end} --deciles 10`
   - Use the fixed interval `{bt_start}` through `{bt_end}` for any recorded candidate.
   - The engine enforces a `{bt_time_limit}`-minute safety limit for an individual backtest.
   - You may use a shorter interval with `--no-save` for a runtime probe. Such a probe is not a candidate result; a recorded version must use the fixed interval.
   - Control concurrency and memory use.
4. Read `report.md`, including the validity decision, long-short metrics, decile returns, and plots. Diagnose both the implementation and the economic hypothesis.
5. Append the cycle to `{memory_file}` with the candidate filename, PASS/FAIL result, key metrics, diagnosis, and the next research decision.
6. Select the strongest current candidate that remains temporally valid and aligned with the assigned idea. Copy its complete file set into the delivery submission for this checkpoint.

Within this cycle, you may perform several tool calls, code edits, runtime probes, or backtests when they are needed to diagnose a candidate. Do not end the invocation without a completed recorded backtest and an updated incumbent submission.

# Research principles

- The final factor is intended for investment research. Do not improve historical metrics through formula hacks, future data, or signals unrelated to the assigned idea.
- Interpret the full report. Sharpe is the primary risk-adjusted return measure, while decile ordering and long- and short-side excess returns help diagnose where the factor works or fails.
- When evidence contradicts the current explanation, revise the construction or change direction within the assigned hypothesis. Preserve a stronger prior incumbent if a new candidate degrades the evidence.
- Keep the implementation economically interpretable. A higher backtest metric does not justify hypothesis drift.

# Computational requirements

- Declare only the required data fields using the metadata described in `data_catalog.md`. Loading an entire table wastes the backtest budget.
- Intraday data are expensive. Use the coarsest available channel that supports the hypothesis and request only needed fields.
- Restrict the historical window to the observations required by the construction.
- Cache reusable rolling intermediates through the framework's incremental-computation interface.
- Prefer vectorized NumPy or pandas operations over loops across stocks.
- Use the timing breakdown in the backtest log to diagnose data loading, factor computation, postprocessing, or evaluation bottlenecks.

# Temporal validity and idea alignment

- Access data only through the time-bounded factor interface. Do not read data files directly with `read_parquet`, `read_csv`, `open`, or `glob`; do not import `loader` or `DataLoader`; do not use absolute paths, parent-directory traversal, hard-coded evaluation dates, or operators aligned to future observations.
- The implementation must remain faithful to the assigned idea. Minor and economically interpretable additions are allowed, but do not add independent reversal, momentum, turnover, valuation, or other predictors merely to raise the backtest score. Hypothesis-consistent transformations and postprocessing choices are allowed when they remain economically interpretable.

# Workspace boundary

Use only the files supplied in the benchmark workspace; do not access external files.
- Write the candidate to `factor.py` and research memory to `{memory_file}`.
- Do not modify the framework source files; they define the common evaluation environment.
- Do not modify the shared `postprocess.toml`. If a candidate uses a different postprocessing specification, use a candidate-specific copy.
- The framework locates market and fundamental data. Do not manually locate or load underlying data files.

# Research memory

After completing the cycle, append the following structure to `{memory_file}`:

```
## Iteration N
- Factor: factor.py
- Decision: PASS / FAIL
- Key metrics: long-short Sharpe / monotonicity / long-side excess / short-side excess
- Analysis: what the evidence supports and what failed (2-4 sentences)
- Next decision: the most useful next experiment or reason to retain the incumbent (1-2 sentences)
```

The memory is the persistent research state for any later invocation. Record failed experiments and negative evidence, not only successful candidates.

# Factor interface

- Store the current factor implementation in `factor.py`.
- Implement the factor entry point described in `factor_base.py`, returning the signal cross-section at T as a pandas Series indexed by stock code.
- Declare the metadata required by `factor_base.py`; the factor name must match its submission identifier.
- Daily, point-in-time fundamental, and intraday interfaces are documented in `data_catalog.md`.
- The runtime-injected catalog below lists the available datasets and fields:

{data_block}

# Postprocessing

The shared pipeline can winsorize, standardize, neutralize industry or style exposures, and transform cross-sectional ranks. Keep transformations intrinsic to the factor definition in the factor computation; configure general cross-sectional postprocessing through a candidate-local copy of `postprocess.toml`.

To test a candidate-specific configuration:

```bash
python backtest.py --factor <name> --start {bt_start} --end {bt_end} --deciles 10 \
  --postprocess-config postprocess.toml
```

The backtest records the actual postprocessing file with the candidate artifact.

# Incumbent submission

Before ending this invocation, include the following filenames in the delivery submission:

```
factor.py
postprocess.toml
```

Choose the incumbent using the feedback, temporal validity, hypothesis alignment, and robustness of the observed performance. If a candidate has a higher Sharpe but shows leakage, unstable concentration, or hypothesis drift, retain a better-supported alternative and record the reason in `{memory_file}`.

Confirm that the recorded candidate has `report.md`, that `{memory_file}` contains the current cycle, and that the delivery submission contains the complete file set. Then reply briefly with the result of this research cycle.
\end{PromptBlock}

\subsection{External Capability Indicator}\label{external-capability-indicator}

\hyperref[tab:capability-index]{Table D.2} records the exact model release and reasoning effort from the September 25, 2026 AAII v4.3.2 snapshot. Doubao Seed 2.1 Pro has no matching entry and is excluded from the capability correlations and empirical fit. We standardize the remaining eight values before fitting the relation in \hyperref[an-empirical-search-scaling-law]{Appendix D.4}.

\begin{table}[htbp]
\caption{External capability values used in the empirical scaling fit.}
\label{tab:capability-index}
\centering
\small
\setlength{\tabcolsep}{3pt}
\begin{tabular}{lrr}
\toprule
Model & AAII & Reasoning Effort \\
\midrule
Claude Opus 5 & 48 & high \\
GLM-5.3 & 45 & max \\
DeepSeek-V4.1-Flash & 39 & max \\
DeepSeek-V4-Pro & 36 & max \\
GLM-5.2 & 34 & max \\
MiniMax-M3 & 29 & reasoning \\
Qwen3.7-Plus & 25 & reasoning \\
DeepSeek-V3.1 & 13 & reasoning \\
Doubao Seed 2.1 Pro & --- & not available \\
\bottomrule
\end{tabular}
\end{table}

\subsection{An Empirical Search Scaling Law}\label{an-empirical-search-scaling-law}

We combine the external capability index with mean net Sharpe ratio at the recorded search depths. Eight systems have matching index values, yielding 24 model--budget observations; Doubao Seed 2.1 Pro is excluded because no matching AAII entry is available. Let \(C_m^*=(C_m-33.63)/10.56\) standardize the index across these systems. Fitting the family from \hyperref[search-scaling-in-autonomous-quantitative-factor-mining]{Section 3} with shared search exponent and \(b_0=0\) gives \[
\widehat{Q}(b,C_m^*)=26.383+0.430C_m^*-25.588b^{-0.020}.
\] This relation explains \(92.7\%\) of the variation in the observed model means, with a root-mean-square error of \(0.138\) Sharpe. The capability term primarily shifts the quality level; iteration depth contributes a common, diminishing gain. Allowing the remaining-gain term to depend on capability reduces the fitting error to \(0.136\). The fitted exponent is \(0.020\), so the power-law curve is close to logarithmic over the observed depths; a log-budget form attains the same in-sample error of \(0.138\).

To test cross-model prediction, we perform leave-one-model-out validation. Each fold fits seven models and predicts the three checkpoints of the eighth. Across eight folds, the aggregate held-out RMSE is \(0.193\) Sharpe for both the power-law and log-budget forms. The common signal across these fits is a diminishing budget-performance curve. These results align with the rank analysis in \hyperref[search-scaling-across-models-and-budgets]{Section 4.2}: base capability strongly orders low-budget quality, whereas the size of an additional search gain is less ordered by capability.

The relation puts the results in one quantitative frame. Models begin at different quality levels, subsequent experiments can narrow a performance gap, and the gain per additional iteration decreases across the measured range. \hyperref[fig:cost-performance]{Figure 3} complements this depth analysis by tracing how each model moves through cost--quality space across all three recorded budgets.

\section{Additional Results}\label{additional-results}

\subsection{Task-Level Metrics}\label{task-level-metrics}

\hyperref[tab:opus-tasks]{Tables E.1}, \hyperref[tab:glm53-tasks]{E.2}, \hyperref[tab:deepseek-v41-flash-tasks]{E.3}, \hyperref[tab:deepseek-v4-pro-tasks]{E.4}, \hyperref[tab:doubao-seed-21-pro-tasks]{E.5}, \hyperref[tab:glm52-tasks]{E.6}, \hyperref[tab:minimax-m3-tasks]{E.7}, \hyperref[tab:qwen37-plus-tasks]{E.8}, and \hyperref[tab:deepseek-v31-tasks]{E.9} report Sharpe, monotonicity, and alignment for all 50 tasks at iteration budgets \(b=3\), \(b=6\), and \(b=10\), with one table per model. For task-level reporting, we select one of the five repeated runs for each model and report its checkpoint metrics consistently across the three budgets.

\clearpage
\begingroup
\footnotesize
\setlength{\tabcolsep}{3.2pt}
\renewcommand{\arraystretch}{0.88}
\setlength\LTleft{0pt}
\setlength\LTright{0pt}
\begin{longtable}{@{}r>{\raggedright\arraybackslash}p{1.82in}rrrrrrrrr@{}}
\caption{Task-level metrics for Claude Opus 5 at iterations 3, 6, and 10.}\label{tab:opus-tasks}\\
\toprule
No. & Task ID & \multicolumn{3}{c}{Sharpe} & \multicolumn{3}{c}{Monotonicity} & \multicolumn{3}{c}{Alignment} \\
\cmidrule(lr){3-5} \cmidrule(lr){6-8} \cmidrule(lr){9-11}
 & & b=3 & b=6 & b=10 & b=3 & b=6 & b=10 & b=3 & b=6 & b=10 \\
\midrule
\endfirsthead
\toprule
No. & Task ID & \multicolumn{3}{c}{Sharpe} & \multicolumn{3}{c}{Monotonicity} & \multicolumn{3}{c}{Alignment} \\
\cmidrule(lr){3-5} \cmidrule(lr){6-8} \cmidrule(lr){9-11}
 & & b=3 & b=6 & b=10 & b=3 & b=6 & b=10 & b=3 & b=6 & b=10 \\
\midrule
\endhead
\bottomrule
\endfoot
1 & \texttt{accr\_excess\_gap} & 2.688 & 2.688 & 2.736 & 1.000 & 1.000 & 0.952 & 4 & 4 & 4 \\
2 & \texttt{anchor\_fit\_value} & 1.791 & 2.170 & 2.547 & 1.000 & 1.000 & 1.000 & 4 & 4 & 3 \\
3 & \texttt{aog\_vwap30\_deq80} & 1.538 & 1.538 & 1.538 & 0.939 & 0.939 & 0.903 & 4 & 4 & 3 \\
4 & \texttt{cohort\_excess\_pulse\_rev} & 2.461 & 2.461 & 3.905 & 1.000 & 1.000 & 1.000 & 4 & 4 & 3 \\
5 & \texttt{droa\_early\_pit} & 3.935 & 3.935 & 3.935 & 0.988 & 0.988 & 0.988 & 5 & 5 & 5 \\
6 & \texttt{earnings\_accel\_trend\_cfo} & 1.403 & 1.403 & 1.638 & 0.915 & 0.964 & 0.976 & 4 & 4 & 4 \\
7 & \texttt{ep\_detrend\_anom\_gate} & 1.914 & 2.073 & 2.210 & 0.964 & 0.988 & 0.988 & 4 & 4 & 4 \\
8 & \texttt{exp\_div\_yield\_fwd} & 1.904 & 1.904 & 2.576 & 0.988 & 0.988 & 0.988 & 4 & 4 & 4 \\
\end{longtable}
\endgroup

\clearpage
\begingroup
\footnotesize
\setlength{\tabcolsep}{3.2pt}
\renewcommand{\arraystretch}{0.88}
\setlength\LTleft{0pt}
\setlength\LTright{0pt}
\begin{longtable}{@{}r>{\raggedright\arraybackslash}p{1.82in}rrrrrrrrr@{}}
\caption{Task-level metrics for GLM-5.3 at iterations 3, 6, and 10.}\label{tab:glm53-tasks}\\
\toprule
No. & Task ID & \multicolumn{3}{c}{Sharpe} & \multicolumn{3}{c}{Monotonicity} & \multicolumn{3}{c}{Alignment} \\
\cmidrule(lr){3-5} \cmidrule(lr){6-8} \cmidrule(lr){9-11}
 & & b=3 & b=6 & b=10 & b=3 & b=6 & b=10 & b=3 & b=6 & b=10 \\
\midrule
\endfirsthead
\toprule
No. & Task ID & \multicolumn{3}{c}{Sharpe} & \multicolumn{3}{c}{Monotonicity} & \multicolumn{3}{c}{Alignment} \\
\cmidrule(lr){3-5} \cmidrule(lr){6-8} \cmidrule(lr){9-11}
 & & b=3 & b=6 & b=10 & b=3 & b=6 & b=10 & b=3 & b=6 & b=10 \\
\midrule
\endhead
\bottomrule
\endfoot
1 & \texttt{accr\_excess\_gap} & 2.516 & 2.516 & 2.622 & 0.952 & 0.952 & 0.952 & 4 & 4 & 5 \\
2 & \texttt{anchor\_fit\_value} & 2.546 & 3.623 & 3.623 & 0.988 & 0.988 & 0.988 & 5 & 5 & 4 \\
3 & \texttt{aog\_vwap30\_deq80} & 0.913 & 1.700 & 2.353 & 0.891 & 0.988 & 1.000 & 4 & 4 & 4 \\
4 & \texttt{cohort\_excess\_pulse\_rev} & 1.387 & 2.653 & 3.034 & 1.000 & 1.000 & 1.000 & 4 & 4 & 4 \\
5 & \texttt{droa\_early\_pit} & 2.790 & 4.291 & 4.341 & 1.000 & 1.000 & 0.988 & 4 & 4 & 4 \\
6 & \texttt{earnings\_accel\_trend\_cfo} & 2.781 & 3.625 & 4.795 & 0.988 & 1.000 & 1.000 & 2 & 2 & 2 \\
7 & \texttt{ep\_detrend\_anom\_gate} & 1.522 & 2.812 & 3.139 & 0.976 & 0.988 & 0.988 & 4 & 4 & 4 \\
8 & \texttt{exp\_div\_yield\_fwd} & 1.610 & 2.216 & 2.216 & 0.988 & 0.988 & 0.964 & 5 & 5 & 4 \\
\end{longtable}
\endgroup

\clearpage
\begingroup
\footnotesize
\setlength{\tabcolsep}{3.2pt}
\renewcommand{\arraystretch}{0.88}
\setlength\LTleft{0pt}
\setlength\LTright{0pt}
\begin{longtable}{@{}r>{\raggedright\arraybackslash}p{1.82in}rrrrrrrrr@{}}
\caption{Task-level metrics for DeepSeek-V4.1-Flash at iterations 3, 6, and 10.}\label{tab:deepseek-v41-flash-tasks}\\
\toprule
No. & Task ID & \multicolumn{3}{c}{Sharpe} & \multicolumn{3}{c}{Monotonicity} & \multicolumn{3}{c}{Alignment} \\
\cmidrule(lr){3-5} \cmidrule(lr){6-8} \cmidrule(lr){9-11}
 & & b=3 & b=6 & b=10 & b=3 & b=6 & b=10 & b=3 & b=6 & b=10 \\
\midrule
\endfirsthead
\toprule
No. & Task ID & \multicolumn{3}{c}{Sharpe} & \multicolumn{3}{c}{Monotonicity} & \multicolumn{3}{c}{Alignment} \\
\cmidrule(lr){3-5} \cmidrule(lr){6-8} \cmidrule(lr){9-11}
 & & b=3 & b=6 & b=10 & b=3 & b=6 & b=10 & b=3 & b=6 & b=10 \\
\midrule
\endhead
\bottomrule
\endfoot
1 & \texttt{accr\_excess\_gap} & 2.298 & 2.298 & 2.298 & 0.964 & 0.964 & 0.964 & 4 & 4 & 4 \\
2 & \texttt{anchor\_fit\_value} & 2.084 & 3.124 & 3.124 & 1.000 & 1.000 & 0.988 & 4 & 4 & 3 \\
3 & \texttt{aog\_vwap30\_deq80} & 0.789 & 0.789 & 1.112 & 0.964 & 0.964 & 0.988 & 4 & 4 & 4 \\
4 & \texttt{cohort\_excess\_pulse\_rev} & 1.397 & 1.397 & 2.753 & 1.000 & 1.000 & 1.000 & 4 & 4 & 3 \\
5 & \texttt{droa\_early\_pit} & 4.162 & 5.063 & 5.063 & 0.988 & 1.000 & 1.000 & 4 & 4 & 4 \\
6 & \texttt{earnings\_accel\_trend\_cfo} & 1.962 & 1.962 & 2.839 & 0.988 & 0.988 & 1.000 & 4 & 4 & 4 \\
7 & \texttt{ep\_detrend\_anom\_gate} & 1.746 & 1.746 & 1.746 & 1.000 & 1.000 & 1.000 & 4 & 4 & 4 \\
8 & \texttt{exp\_div\_yield\_fwd} & 2.542 & 2.580 & 2.580 & 1.000 & 1.000 & 1.000 & 4 & 4 & 5 \\
\end{longtable}
\endgroup

\clearpage
\begingroup
\footnotesize
\setlength{\tabcolsep}{3.2pt}
\renewcommand{\arraystretch}{0.88}
\setlength\LTleft{0pt}
\setlength\LTright{0pt}
\begin{longtable}{@{}r>{\raggedright\arraybackslash}p{1.82in}rrrrrrrrr@{}}
\caption{Task-level metrics for DeepSeek-V4-Pro at iterations 3, 6, and 10.}\label{tab:deepseek-v4-pro-tasks}\\
\toprule
No. & Task ID & \multicolumn{3}{c}{Sharpe} & \multicolumn{3}{c}{Monotonicity} & \multicolumn{3}{c}{Alignment} \\
\cmidrule(lr){3-5} \cmidrule(lr){6-8} \cmidrule(lr){9-11}
 & & b=3 & b=6 & b=10 & b=3 & b=6 & b=10 & b=3 & b=6 & b=10 \\
\midrule
\endfirsthead
\toprule
No. & Task ID & \multicolumn{3}{c}{Sharpe} & \multicolumn{3}{c}{Monotonicity} & \multicolumn{3}{c}{Alignment} \\
\cmidrule(lr){3-5} \cmidrule(lr){6-8} \cmidrule(lr){9-11}
 & & b=3 & b=6 & b=10 & b=3 & b=6 & b=10 & b=3 & b=6 & b=10 \\
\midrule
\endhead
\bottomrule
\endfoot
1 & \texttt{accr\_excess\_gap} & 1.721 & 1.887 & 2.148 & 0.988 & 0.988 & 0.988 & 4 & 4 & 4 \\
2 & \texttt{anchor\_fit\_value} & 2.137 & 2.137 & 1.516 & 1.000 & 1.000 & 1.000 & 4 & 4 & 4 \\
3 & \texttt{aog\_vwap30\_deq80} & 0.601 & 0.601 & 0.149 & 0.842 & 0.842 & 0.333 & 5 & 5 & 2 \\
4 & \texttt{cohort\_excess\_pulse\_rev} & 1.362 & 1.362 & 2.234 & 0.964 & 0.964 & 1.000 & 4 & 4 & 4 \\
5 & \texttt{droa\_early\_pit} & 1.456 & 1.628 & 4.406 & 0.964 & 1.000 & 1.000 & 5 & 5 & 4 \\
6 & \texttt{earnings\_accel\_trend\_cfo} & 1.194 & 2.934 & 3.026 & 0.915 & 0.915 & 1.000 & 4 & 4 & 2 \\
7 & \texttt{ep\_detrend\_anom\_gate} & 2.239 & 2.239 & 2.611 & 0.988 & 1.000 & 1.000 & 4 & 4 & 5 \\
8 & \texttt{exp\_div\_yield\_fwd} & 1.761 & 1.761 & 1.860 & 0.964 & 0.964 & 0.964 & 5 & 5 & 5 \\
\end{longtable}
\endgroup

\clearpage
\begingroup
\footnotesize
\setlength{\tabcolsep}{3.2pt}
\renewcommand{\arraystretch}{0.88}
\setlength\LTleft{0pt}
\setlength\LTright{0pt}
\begin{longtable}{@{}r>{\raggedright\arraybackslash}p{1.82in}rrrrrrrrr@{}}
\caption{Task-level metrics for Doubao Seed 2.1 Pro at iterations 3, 6, and 10.}\label{tab:doubao-seed-21-pro-tasks}\\
\toprule
No. & Task ID & \multicolumn{3}{c}{Sharpe} & \multicolumn{3}{c}{Monotonicity} & \multicolumn{3}{c}{Alignment} \\
\cmidrule(lr){3-5} \cmidrule(lr){6-8} \cmidrule(lr){9-11}
 & & b=3 & b=6 & b=10 & b=3 & b=6 & b=10 & b=3 & b=6 & b=10 \\
\midrule
\endfirsthead
\toprule
No. & Task ID & \multicolumn{3}{c}{Sharpe} & \multicolumn{3}{c}{Monotonicity} & \multicolumn{3}{c}{Alignment} \\
\cmidrule(lr){3-5} \cmidrule(lr){6-8} \cmidrule(lr){9-11}
 & & b=3 & b=6 & b=10 & b=3 & b=6 & b=10 & b=3 & b=6 & b=10 \\
\midrule
\endhead
\bottomrule
\endfoot
1 & \texttt{accr\_excess\_gap} & 1.953 & 2.313 & 2.340 & 0.952 & 0.964 & 0.964 & 4 & 4 & 3 \\
2 & \texttt{anchor\_fit\_value} & 1.333 & 1.629 & 2.015 & 1.000 & 1.000 & 1.000 & 4 & 4 & 3 \\
3 & \texttt{aog\_vwap30\_deq80} & 3.087 & 3.090 & 3.235 & 1.000 & 1.000 & 1.000 & 5 & 5 & 4 \\
4 & \texttt{cohort\_excess\_pulse\_rev} & 1.278 & 1.816 & 2.074 & 0.976 & 1.000 & 1.000 & 5 & 5 & 4 \\
5 & \texttt{droa\_early\_pit} & 3.026 & 3.192 & 3.361 & 1.000 & 1.000 & 1.000 & 3 & 3 & 4 \\
6 & \texttt{earnings\_accel\_trend\_cfo} & 2.384 & 2.434 & 3.248 & 1.000 & 1.000 & 1.000 & 3 & 3 & 3 \\
7 & \texttt{ep\_detrend\_anom\_gate} & 1.970 & 2.221 & 2.267 & 1.000 & 1.000 & 0.988 & 3 & 3 & 4 \\
8 & \texttt{exp\_div\_yield\_fwd} & 1.739 & 1.790 & 1.890 & 0.964 & 0.964 & 0.964 & 5 & 5 & 3 \\
\end{longtable}
\endgroup

\clearpage
\begingroup
\footnotesize
\setlength{\tabcolsep}{3.2pt}
\renewcommand{\arraystretch}{0.88}
\setlength\LTleft{0pt}
\setlength\LTright{0pt}
\begin{longtable}{@{}r>{\raggedright\arraybackslash}p{1.82in}rrrrrrrrr@{}}
\caption{Task-level metrics for GLM-5.2 at iterations 3, 6, and 10.}\label{tab:glm52-tasks}\\
\toprule
No. & Task ID & \multicolumn{3}{c}{Sharpe} & \multicolumn{3}{c}{Monotonicity} & \multicolumn{3}{c}{Alignment} \\
\cmidrule(lr){3-5} \cmidrule(lr){6-8} \cmidrule(lr){9-11}
 & & b=3 & b=6 & b=10 & b=3 & b=6 & b=10 & b=3 & b=6 & b=10 \\
\midrule
\endfirsthead
\toprule
No. & Task ID & \multicolumn{3}{c}{Sharpe} & \multicolumn{3}{c}{Monotonicity} & \multicolumn{3}{c}{Alignment} \\
\cmidrule(lr){3-5} \cmidrule(lr){6-8} \cmidrule(lr){9-11}
 & & b=3 & b=6 & b=10 & b=3 & b=6 & b=10 & b=3 & b=6 & b=10 \\
\midrule
\endhead
\bottomrule
\endfoot
1 & \texttt{accr\_excess\_gap} & 2.168 & 2.168 & 1.266 & 0.976 & 0.976 & 0.976 & 4 & 4 & 4 \\
2 & \texttt{anchor\_fit\_value} & 1.014 & 1.385 & 1.588 & 0.988 & 1.000 & 1.000 & 4 & 4 & 4 \\
3 & \texttt{aog\_vwap30\_deq80} & 1.609 & 1.609 & 0.670 & 0.927 & 0.927 & 0.939 & 4 & 4 & 4 \\
4 & \texttt{cohort\_excess\_pulse\_rev} & 1.448 & 1.448 & 1.519 & 0.988 & 0.988 & 1.000 & 4 & 4 & 4 \\
5 & \texttt{droa\_early\_pit} & 2.095 & 2.095 & 2.821 & 1.000 & 1.000 & 0.976 & 3 & 3 & 3 \\
6 & \texttt{earnings\_accel\_trend\_cfo} & 1.808 & 2.110 & 3.030 & 0.976 & 0.988 & 1.000 & 4 & 4 & 2 \\
7 & \texttt{ep\_detrend\_anom\_gate} & 1.768 & 2.246 & 2.246 & 1.000 & 1.000 & 1.000 & 4 & 4 & 4 \\
8 & \texttt{exp\_div\_yield\_fwd} & 2.095 & 2.095 & 1.884 & 1.000 & 1.000 & 1.000 & 4 & 4 & 3 \\
\end{longtable}
\endgroup

\clearpage
\begingroup
\footnotesize
\setlength{\tabcolsep}{3.2pt}
\renewcommand{\arraystretch}{0.88}
\setlength\LTleft{0pt}
\setlength\LTright{0pt}
\begin{longtable}{@{}r>{\raggedright\arraybackslash}p{1.82in}rrrrrrrrr@{}}
\caption{Task-level metrics for MiniMax-M3 at iterations 3, 6, and 10.}\label{tab:minimax-m3-tasks}\\
\toprule
No. & Task ID & \multicolumn{3}{c}{Sharpe} & \multicolumn{3}{c}{Monotonicity} & \multicolumn{3}{c}{Alignment} \\
\cmidrule(lr){3-5} \cmidrule(lr){6-8} \cmidrule(lr){9-11}
 & & b=3 & b=6 & b=10 & b=3 & b=6 & b=10 & b=3 & b=6 & b=10 \\
\midrule
\endfirsthead
\toprule
No. & Task ID & \multicolumn{3}{c}{Sharpe} & \multicolumn{3}{c}{Monotonicity} & \multicolumn{3}{c}{Alignment} \\
\cmidrule(lr){3-5} \cmidrule(lr){6-8} \cmidrule(lr){9-11}
 & & b=3 & b=6 & b=10 & b=3 & b=6 & b=10 & b=3 & b=6 & b=10 \\
\midrule
\endhead
\bottomrule
\endfoot
1 & \texttt{accr\_excess\_gap} & 1.256 & 1.262 & 1.423 & 0.915 & 0.976 & 0.976 & 5 & 2 & 2 \\
2 & \texttt{anchor\_fit\_value} & 1.944 & 1.993 & 1.772 & 1.000 & 1.000 & 1.000 & 4 & 4 & 4 \\
3 & \texttt{aog\_vwap30\_deq80} & -0.035 & -3.320 & -0.265 & 0.745 & 0.879 & 0.915 & 5 & 4 & 4 \\
4 & \texttt{cohort\_excess\_pulse\_rev} & 1.267 & 1.431 & 1.398 & 0.976 & 0.964 & 0.988 & 4 & 5 & 4 \\
5 & \texttt{droa\_early\_pit} & 0.887 & 0.810 & 1.148 & 0.976 & 0.988 & 0.939 & 5 & 3 & 4 \\
6 & \texttt{earnings\_accel\_trend\_cfo} & 3.514 & 2.541 & 3.680 & 0.988 & 0.976 & 0.988 & 3 & 3 & 4 \\
7 & \texttt{ep\_detrend\_anom\_gate} & -1.171 & -0.231 & -0.454 & 0.915 & 0.976 & 0.964 & 4 & 5 & 4 \\
8 & \texttt{exp\_div\_yield\_fwd} & 2.089 & 1.221 & 2.161 & 0.964 & 1.000 & 0.952 & 3 & 2 & 4 \\
\end{longtable}
\endgroup

\clearpage
\begingroup
\footnotesize
\setlength{\tabcolsep}{3.2pt}
\renewcommand{\arraystretch}{0.88}
\setlength\LTleft{0pt}
\setlength\LTright{0pt}
\begin{longtable}{@{}r>{\raggedright\arraybackslash}p{1.82in}rrrrrrrrr@{}}
\caption{Task-level metrics for Qwen3.7-Plus at iterations 3, 6, and 10.}\label{tab:qwen37-plus-tasks}\\
\toprule
No. & Task ID & \multicolumn{3}{c}{Sharpe} & \multicolumn{3}{c}{Monotonicity} & \multicolumn{3}{c}{Alignment} \\
\cmidrule(lr){3-5} \cmidrule(lr){6-8} \cmidrule(lr){9-11}
 & & b=3 & b=6 & b=10 & b=3 & b=6 & b=10 & b=3 & b=6 & b=10 \\
\midrule
\endfirsthead
\toprule
No. & Task ID & \multicolumn{3}{c}{Sharpe} & \multicolumn{3}{c}{Monotonicity} & \multicolumn{3}{c}{Alignment} \\
\cmidrule(lr){3-5} \cmidrule(lr){6-8} \cmidrule(lr){9-11}
 & & b=3 & b=6 & b=10 & b=3 & b=6 & b=10 & b=3 & b=6 & b=10 \\
\midrule
\endhead
\bottomrule
\endfoot
1 & \texttt{accr\_excess\_gap} & 0.016 & 0.745 & 2.615 & 0.491 & 0.539 & 1.000 & 3 & 3 & 0 \\
2 & \texttt{anchor\_fit\_value} & 1.502 & 1.502 & 1.689 & 0.988 & 0.988 & 0.988 & 2 & 2 & 3 \\
3 & \texttt{aog\_vwap30\_deq80} & 0.710 & 0.710 & 0.710 & 0.842 & 0.842 & 0.842 & 4 & 4 & 4 \\
4 & \texttt{cohort\_excess\_pulse\_rev} & 1.658 & 1.658 & 2.404 & 0.988 & 1.000 & 1.000 & 4 & 4 & 3 \\
5 & \texttt{droa\_early\_pit} & 1.865 & 2.543 & 2.851 & 0.964 & 0.988 & 1.000 & 3 & 3 & 2 \\
6 & \texttt{earnings\_accel\_trend\_cfo} & -0.319 & 3.703 & 4.172 & 0.612 & 1.000 & 1.000 & 4 & 4 & 1 \\
7 & \texttt{ep\_detrend\_anom\_gate} & 0.336 & 0.336 & 1.718 & 0.818 & 0.818 & 1.000 & 4 & 4 & 1 \\
8 & \texttt{exp\_div\_yield\_fwd} & 1.633 & 1.964 & 1.964 & 0.964 & 0.988 & 0.988 & 3 & 3 & 1 \\
\end{longtable}
\endgroup

\clearpage
\begingroup
\footnotesize
\setlength{\tabcolsep}{3.2pt}
\renewcommand{\arraystretch}{0.88}
\setlength\LTleft{0pt}
\setlength\LTright{0pt}
\begin{longtable}{@{}r>{\raggedright\arraybackslash}p{1.82in}rrrrrrrrr@{}}
\caption{Task-level metrics for DeepSeek-V3.1 at iterations 3, 6, and 10.}\label{tab:deepseek-v31-tasks}\\
\toprule
No. & Task ID & \multicolumn{3}{c}{Sharpe} & \multicolumn{3}{c}{Monotonicity} & \multicolumn{3}{c}{Alignment} \\
\cmidrule(lr){3-5} \cmidrule(lr){6-8} \cmidrule(lr){9-11}
 & & b=3 & b=6 & b=10 & b=3 & b=6 & b=10 & b=3 & b=6 & b=10 \\
\midrule
\endfirsthead
\toprule
No. & Task ID & \multicolumn{3}{c}{Sharpe} & \multicolumn{3}{c}{Monotonicity} & \multicolumn{3}{c}{Alignment} \\
\cmidrule(lr){3-5} \cmidrule(lr){6-8} \cmidrule(lr){9-11}
 & & b=3 & b=6 & b=10 & b=3 & b=6 & b=10 & b=3 & b=6 & b=10 \\
\midrule
\endhead
\bottomrule
\endfoot
1 & \texttt{accr\_excess\_gap} & -0.700 & -0.524 & -0.524 & 0.250 & 0.273 & 0.273 & 4 & 4 & 3 \\
2 & \texttt{anchor\_fit\_value} & 1.741 & 1.741 & -1.628 & 1.000 & 1.000 & 0.661 & 3 & 3 & 4 \\
3 & \texttt{aog\_vwap30\_deq80} & 0.334 & 0.334 & -0.092 & 0.709 & 0.915 & 1.000 & 5 & 5 & 2 \\
4 & \texttt{cohort\_excess\_pulse\_rev} & 1.360 & 1.360 & 1.744 & 0.976 & 0.976 & 1.000 & 5 & 5 & 4 \\
5 & \texttt{droa\_early\_pit} & 1.548 & 1.548 & 4.489 & 0.988 & 0.988 & 1.000 & 3 & 3 & 4 \\
6 & \texttt{earnings\_accel\_trend\_cfo} & 1.716 & 1.848 & 1.993 & 0.976 & 0.976 & 0.988 & 4 & 4 & 4 \\
7 & \texttt{ep\_detrend\_anom\_gate} & -0.099 & -0.029 & 0.010 & 0.855 & 0.855 & 0.855 & 1 & 1 & 4 \\
8 & \texttt{exp\_div\_yield\_fwd} & 0.939 & 2.299 & 2.299 & 0.952 & 0.988 & 0.988 & 4 & 4 & 1 \\
\end{longtable}
\endgroup

\subsection{Research Cases}\small\label{research-cases}

The following cases pair a higher-scoring and a lower-scoring trajectory for each behavioral difference discussed in \hyperref[qualitative-analysis-of-research-trajectories]{Section 4.4}. The quoted passages are English translations of the agents' persistent research memories; metrics and decisions are preserved from the original records.

\subsubsection{Search Redirection after a Plateau}\label{search-redirection-after-a-plateau}

\textbf{Higher-scoring behavior: Claude Opus 5 on \texttt{burst\_support\_imb}.} The model first tested whether a longer aggregation window could jointly reduce turnover and improve measurement precision. Here, \(W\) is the lookback window in trading days, and \(Q\) is the intraday volume-quantile threshold used to identify unusually large trades. When the result showed that lower turnover came with proportional alpha decay, the model stopped searching that axis and moved to structural changes in trading frequency and signal normalization.

\begin{quote}
\textbf{Iteration 2.} Extending \(W\) from 40 to 80 reduced turnover by 31\%, but gross annual return also fell by 20\%; net Sharpe was essentially unchanged at 1.368 versus 1.336. The signal has a real decay horizon. \(W\) is not the way forward, so I will not spend another iteration testing \(W=20\).

\textbf{Iteration 4.} Both \(W\) and \(Q\) are exhausted, and the remaining bottleneck is structural rather than parametric. I will test weekly signal refresh, because the idea describes a persistent behavioral characteristic that need not be traded every day, and replace the bounded ratio with share-normalized signed burst volume. Weekly refresh reduced turnover from 14.6\% to 8.3\% and raised net Sharpe from 1.368 to 1.753; share normalization raised it to 2.565.
\end{quote}

The revision changes what the factor measures and how it is traded. It follows from a diagnosed failure of the current search direction rather than another nearby parameter choice.

\textbf{Lower-scoring behavior: MiniMax-M3 on \texttt{sharp\_minute\_impact\_persist}.} The model correctly identified turnover as a bottleneck, but successive revisions continued to smooth the same impact-persistence representation by enlarging its aggregation window. Here, \(W\) again denotes the lookback window in trading days.

\begin{quote}
\textbf{Iteration 4.} Aggregating the numerator and denominator over \(W=40\) days reduces noise, but net Sharpe remains \(-0.275\) with 14.25\% turnover. The signal exists, but it is not strong enough to survive costs.

\textbf{Iterations 5--6.} Increasing \(W\) to 60 and changing the retained intraday observations raises gross Sharpe but lowers net Sharpe to \(-0.365\) and monotonicity to 0.806. Extending \(W\) again to 100 reduces turnover to 7.78\% and makes net Sharpe positive at 0.332, while monotonicity remains 0.806. The next step is to keep working on the window and extreme-observation treatment.
\end{quote}

The later experiments repeatedly trade signal sharpness for lower turnover along the same smoothing axis. Even after monotonicity deteriorates, the next decision remains another window or extreme-value adjustment rather than a new explanation of the factor's weak economic separation.

\subsubsection{Attribution of Experimental Feedback}\label{attribution-of-experimental-feedback}

\textbf{Higher-scoring behavior: Claude Opus 5 on \texttt{accr\_excess\_gap}.} The initial composite was decomposed into controlled single-component variants before weights were revised. In the excerpt, \(z(\cdot)\) denotes cross-sectional standardization, XACC is the main excess-accrual component, \(\Delta\)XACC is its change component, and GAP is the discrepancy between two accounting constructions.

\begin{quote}
\textbf{Iteration 1.} The composite is \(z(-\mathrm{XACC})+0.5z(-\Delta\mathrm{XACC})+0.5z(-\mathrm{GAP})\). The factor is stable but weak. Before changing weights, run full-period ablations for XACC only, \(\Delta\)XACC only, and GAP only to determine whether the auxiliary components add information or dilute the main signal.

\textbf{Iteration 2.} XACC alone has Sharpe 2.362, \(\Delta\)XACC alone 1.604, and GAP alone 0.256 with monotonicity 0.576. GAP is mostly accounting noise and its weight dilutes XACC, although a small weight improves middle-decile ordering. Reduce the GAP weight from 0.50 to 0.15, retain \(\Delta\)XACC at a lower weight, and normalize weights over the components available for each stock.
\end{quote}

The controlled ablation separates a plausible economic story from its measured contribution. The next implementation directly follows the component-level evidence.

\textbf{Lower-scoring behavior: DeepSeek-V3.1 on \texttt{pullup\_granularity\_asym}.} Early rounds attributed weak results to regression, volume weighting, temporal smoothing, and economic sign. A later implementation check revealed that the code had sliced stocks rather than dates, invalidating those interpretations. EWMA denotes an exponentially weighted moving average, and \texttt{WINDOW} is the intended number of historical dates in the aggregation.

\begin{quote}
\textbf{Iterations 1--3.} Regression and volume weighting appear to distort the signal. Removing them restores the expected direction and monotonicity, while EWMA smoothing is catastrophically bad. The direction is exhausted after these alternatives fail.

\textbf{Iteration 5.} A serious bug was found: \texttt{a.iloc{[}-WINDOW:{]}} selected the last 20 stock rows rather than the last 20 date columns. Every earlier version was therefore computed from only 20 stocks and its result is unreliable.

\textbf{Iteration 6.} After correcting the slice and rerunning the factor, the raw signal predicts in the opposite direction from the research hypothesis. What had been interpreted as overreaction is better described in these data as concentrated buying followed by continuation.
\end{quote}

Because the construction error was discovered only after several economic explanations had been formed, the earlier feedback was attributed to mechanisms that the implemented experiment did not actually test.

\subsubsection{Objective-Consistent Candidate Selection}\label{objective-consistent-candidate-selection}

\textbf{Higher-scoring behavior: Claude Opus 5 on \texttt{smart\_turnover\_split\_rev}.} A higher-Sharpe candidate was rejected after a control showed that its gain came from an independent reversal exposure. The zero-sum constraint requires the within-window weights to sum to zero, removing the ordinary window-average return from the constructed signal.

\begin{quote}
\textbf{Iterations 9--10.} The high-participation-only candidate reaches Sharpe 3.866, above the incumbent's 3.361. Removing the zero-sum constraint, however, introduces the negative window-average return, which is an ordinary intraday-reversal factor. The pure-reversal control has Sharpe 2.754, and the high-participation-only candidate's loading on it rises from 0.467 to 0.636. The additional performance therefore comes from a different signal source rather than a better implementation of participation-conditioned return. Deliver the lower-scoring zero-sum incumbent.
\end{quote}

Here final selection preserves the intended mechanism even when the alternative passes the financial threshold by a wider margin.

\textbf{Lower-scoring behavior: MiniMax-M3 on \texttt{order\_mix\_surprise\_amp}.} The selected candidate appends a generic reversal component after the report-grounded order-mix signal plateaus. Here, \(K\) is the historical lookback in trading days, and reversal-40 is a separate 40-day return-reversal signal.

\begin{quote}
\textbf{Iterations 8--9.} The pure \(K=90\) order-mix signal has Sharpe 0.184. Adding a 30\% 40-day reversal component raises Sharpe to 0.855 and produces perfect monotonicity. This is the best candidate in the current run.

\textbf{Final selection.} Deliver the blended version. The construction retains the order-mix surprise and amplitude weighting, with a 30\% reversal-40 component added for complementarity.
\end{quote}

The selection follows the improved backtest even though the added component supplies an independent return-reversal signal. This is the type of metric-driven extension that the alignment audit is designed to distinguish from refinement of the assigned hypothesis.

\end{document}